%% file: main.tex
\documentclass[nonacm,manuscript]{acmart}
\AtBeginDocument{%
  }

\usepackage{footnote} 
\usepackage[para,online,flushleft]{threeparttable}
\usepackage{graphicx}
\usepackage{subcaption}
\usepackage[normalem]{ulem}
\usepackage{bm}
\usepackage{multirow}
\usepackage{pifont} 
\usepackage{wrapfig}

\usepackage{adjustbox}
\usepackage{pgfplots} 
\usepackage{pgf-pie} 

\providecommand{\showCODEN}[1]{}
\usepackage{xcolor}
\usepackage{adjustbox}

\usepackage{mdframed}

\newif\iflatexdiff
\latexdifffalse

\newenvironment{diffbox}
{
\iflatexdiff
\begin{mdframed}[
    linewidth=1pt,
    linecolor=blue,
    backgroundcolor=blue!2,
    skipabove=0.5em,
    skipbelow=0.5em
]
\fi
}
{
\iflatexdiff
\end{mdframed}
\fi
}

\begin{document}

\title{Accelerated Execution of Bayesian Neural Networks using a Single Probabilistic Forward Pass and Code Generation}

\author{Bernhard Klein}
\email{bernhard.klein@ziti.uni-heidelberg.de}
\orcid{0000-0003-0497-5748}
\affiliation{%
  \institution{Heidelberg University}
  \city{Heidelberg}
  \country{Germany}
}

\author{Falk Selker}
\email{falk.selker@neclab.eu}
\orcid{0009-0003-4476-6884}
\affiliation{%
  \institution{Heidelberg University}
  \city{Heidelberg}
  \country{Germany}
}

\author{Hendrik Borras}
\email{hendrik.borras@ziti.uni-heidelberg.de}
\orcid{0000-0002-2411-2416}
\affiliation{%
  \institution{Heidelberg University}
  \city{Heidelberg}
  \country{Germany}
}

\author{Sophie Steger}
\email{sophie.steger@tugraz.at}
\orcid{0000-0001-8766-3277}
\affiliation{%
  \institution{Graz University of Technology}
  \city{Graz}
  \country{Austria}
}

\author{Franz Pernkopf}
\email{pernkopf@tugraz.at}
\orcid{0000-0002-6356-3367}
\affiliation{%
  \institution{Graz University of Technology}
  \city{Graz}
  \country{Austria}
}

\author{Holger Fr{\"o}ning}
\email{holger.froening@ziti.uni-heidelberg.de}
\orcid{0000-0001-9562-0680}
\affiliation{%
  \institution{Heidelberg University}
  \city{Heidelberg}
  \country{Germany}
}

\renewcommand{\shortauthors}{Klein et al.}

\input{abstract}


\maketitle
\input{01-introduction}

\input{02-related-work}

\input{03-pfp-concept}

\input{04-pfp-training}

\input{05-operator-library}

\input{06-tuning}

\input{06a-pBNNs}
\input{07-conclusion}

\input{acknowledgement}

\bibliographystyle{ACM-Reference-Format}
\bibliography{references}
\end{document}

%% file: abstract.tex
\begin{abstract}

Machine learning models excel across applications like medical diagnostics, weather forecasting, natural language processing and autonomous driving, yet their inadequate handling of uncertainty remains crucial for safety-critical applications.
Traditional neural networks fail to recognize out-of-domain (OOD) data, often producing incorrect predictions without indicating uncertainty.
Bayesian neural networks (BNNs) offer a principled framework for uncertainty estimation by providing statistically grounded probabilities alongside predictions.

Despite these advantages, BNNs suffer from high computational costs in both training and prediction.
Predictions require sampling over weight distributions and multiple forward passes.
To address this challenge, the Probabilistic Forward Pass (PFP) serves as an extreme approximation of Stochastic Variational Inference (SVI).
While SVI assumes Gaussian-distributed weights without restrictions on activations, PFP extends this assumption to activations.  
This enables a fully analytical uncertainty propagation, replacing multiple forward passes with a single, more complex forward pass operating on probability distributions.

Thus, PFP requires specialized Gaussian-propagating operators absent from standard deep learning libraries.  
We present an end-to-end pipeline for training, compilation, optimization, and deployment of PFP-based BNNs on embedded ARM CPUs.  
By implementing custom operators in the deep learning compiler TVM, we develop a dedicated operator library for multilayer perceptrons and convolutional neural networks.  
Multiple operator implementations, along with manual and automatic tuning techniques, are applied to maximize efficiency.  
Ablation studies show that PFP consistently outperforms SVI in computational efficiency.  
For small mini-batch sizes, critical for low-latency applications, our approach achieves speedups of up to $3900\times$.  

Our results show that PFP-based BNNs achieve performance comparable to SVI-BNNs for uncertainty disentanglement on the Dirty-MNIST dataset and for complex OOD detection on CIFAR-10 against SVHN, all while significantly reducing computational overhead.
These findings underscore the potential of combining Bayesian approximations with code generation and operator tuning to accelerate BNN predictions and enable their efficient deployment on resource-constrained embedded systems.

\end{abstract}

%% file: 01-introduction.tex
\section{Introduction}
\label{sec:intro}

Machine learning (ML) based on deep (artificial) neural networks (DNNs) has revolutionized reasoning under uncertainty, and contributed to groundbreaking improvements in various prediction tasks including image, speech and signal processing.
While it comes at tremendous cost, its excellent performance outweighs this downside, and various approaches are being pursued to bridge the gap between resource requirements and hardware capabilities.
Specialized processor architectures are a notable example, which can substantially accelerate the execution of DNNs; however, they come at the cost of generality, often supporting only a limited set of operators, while at the same time, innovation in ML makes fast progress.
As a result, the community has put a remarkable effort to the development of ML compilers, which map operators to special hardware primitives, for instance machine instructions for matrix-matrix multiply. 
This mapping involves several steps of considerable complexity, including loop and memory transformations and specifying which computations should be implemented with which machine instructions.
Recent interest from the hardware and compilation community also extends to code generation for ML~\cite{Chen2018tvm,Lattner2021mlir}.  
This includes efforts in auto-tuning to identify operator implementations optimized for specific hardware architectures~\cite{Roesch2018relay,Lai2023relax,Shao2022tvmMetaScheduler}.

While DNNs are among the most effective methods for learning and predicting under uncertainty, they fundamentally lack the ability to quantify any uncertainty in a principled, mathematical manner.  
In classification tasks, for example, the softmax function is commonly used to produce class-wise output scores that are interpreted as probabilities.  
However, these softmax-derived values do not possess a rigorous probabilistic interpretation; they are best understood as confidence scores without a theoretical foundation grounded in probability theory.  
This limitation becomes particularly apparent when models are exposed to data samples that differ significantly from the training distribution.  
These are commonly referred to as \emph{out-of-domain (OOD)} data.  
In such cases, neural networks often produce highly confident yet incorrect predictions, without any mechanism to signal elevated uncertainty or to indicate that the input lies outside the model's domain of competence.  

A theoretical framework to allow neural architectures to say "I don't know" is cast by Bayesian statistics, resulting in a class of probabilistic models called Bayesian neural networks (BNNs).
They provide predictions in the form of a formal probability distribution, in which, in a simple setting, the mean would correspond to the prediction itself while the uncertainty is coded as variance or standard deviation of this distribution.

BNNs, while providing a mathematically rigorous framework for uncertainty estimation, incur an even greater computational burden than standard DNNs, which are already exceedingly expensive in both training and inference.
Their high computational cost for predictions arises from the need to sample weights from posterior distributions and perform multiple forward passes to estimate predictive uncertainty.
These repeated sampling and forward pass operations during prediction significantly increase latency, posing challenges for deploying BNNs in resource-constrained environments such as Internet of Things (IoT) devices and embedded systems.

In general, probabilistic modeling has gained significant interest within the ML community, leading to extensive related work~\cite{ghahramani2015probabilisticML,murphy2012probabilisticPerspective}.
However, relatively little research addresses probabilistic modeling from a hardware perspective.
Notable exceptions exist regarding hardware-efficient probabilistic circuits~\cite{Leslin2022hwPc,yao2024HwPc}.
Furthermore, a hardware-aware cost metric for probabilistic models has been introduced within tractable learning~\cite{Olascoaga2019HwProb}.
Specialized hardware accelerators for BNNs employing numerous on-chip pseudo-random number generators have also been proposed~\cite{shiftbnn,cai2018,brckerhoffplckelmann2024probabilistic}.

Probabilistic inference based on BNNs is a problem of exceptional complexity due to the inherent complexity of random variable algebra.
At the same time, uncertainty is of high importance when bringing ML models into the "wild", where sensors are hindered by noise, occlusions, and previously unseen settings.\footnote{In San Francisco 2023 a driverless car just got stuck in wet concrete, since it was unaware of the situation that wet concrete exists. \url{https://www.nytimes.com/2023/08/17/us/driverless-car-accident-sf.html}}
To successfully deploy BNNs on resource-constrained devices requires to combine multiple methods to bridge the gap among BNN requirements on memory, compute and the corresponding hardware capabilities.

\begin{figure}
    \centering
	\begin{subfigure}[t]{0.55\textwidth}
    	\includegraphics[width=\textwidth]{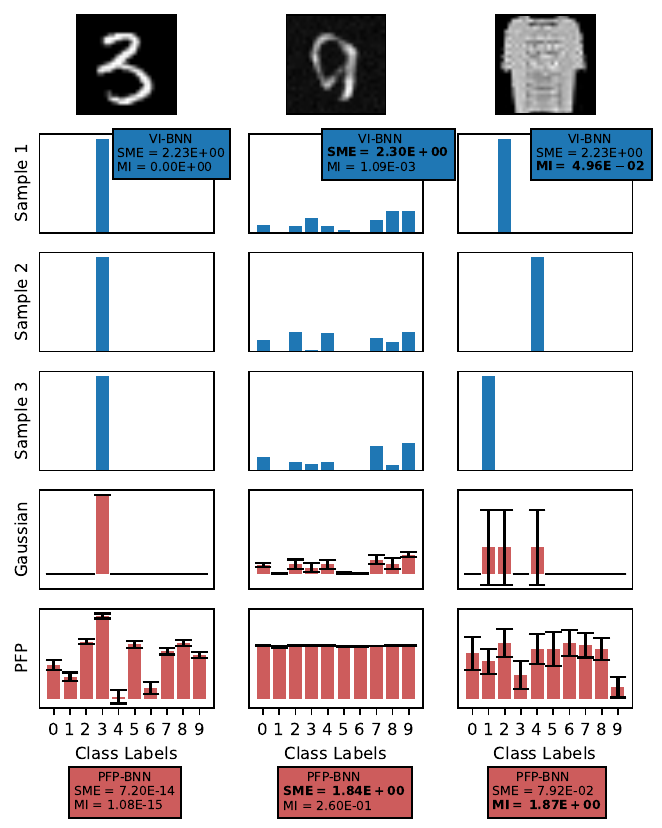}    
		\caption{Illustration of BNN Predictions}
    	\label{fig:example_samples}
	\end{subfigure}
	\hspace{1em}
	\begin{subfigure}[t]{0.4\textwidth}
    	\includegraphics[width=\textwidth]{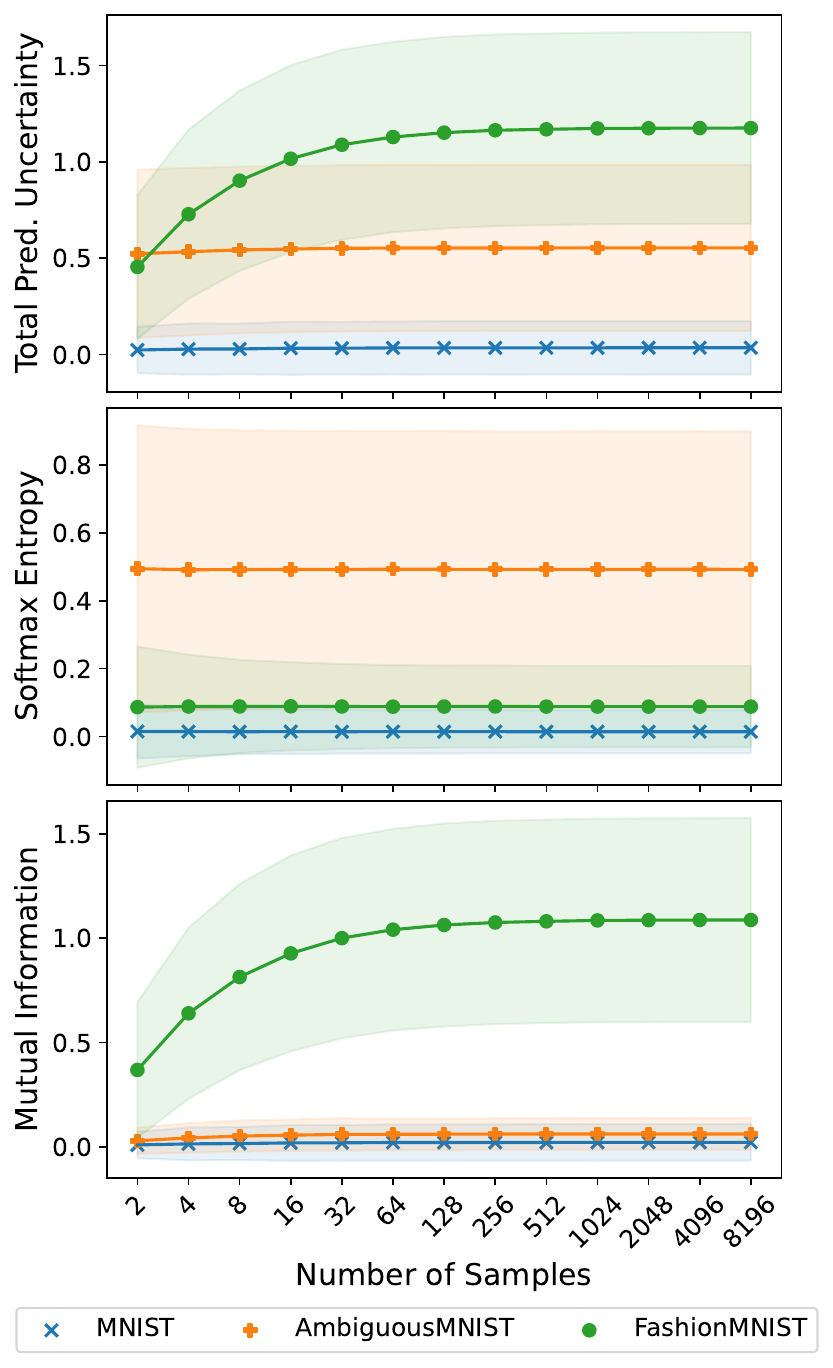}
		\caption{Effect of Number of Samples in SVI-BNNs}
    	\label{fig:number_of_samples_vi}
	\end{subfigure}
	\caption{(\subref{fig:example_samples}) Exemplary predictions from a SVI-BNN (blue), its Gaussian representation, and PFP~\cite{roth2016pfp}, showcasing MNIST~\cite{Lecun1998mnist} and Ambiguous-MNIST~\cite{Mukhoti2022dirtyMNIST} as in-domain examples, and Fashion-MNIST~\cite{xiao2017fashionMnist} as an out-of-domain sample. Variability in class predictions demonstrates aleatoric uncertainty (higher Softmax Entropy, SME), whereas variability across predictive samples indicates epistemic uncertainty (higher Mutual Information, MI). SVI and PFP effectively quantify these uncertainties. Note: Only three SVI samples shown, insufficient for robust estimation. (\subref{fig:number_of_samples_vi}) Influence of predictive sample count on uncertainty metrics. Softmax Entropy (aleatoric uncertainty) remains stable, while Total Predictive Uncertainty and Mutual Information (epistemic uncertainty), especially for out-of-domain data (Fashion-MNIST), require more samples for reliable OOD detection.}
    \label{fig:example_samples_and_number_of_samples_vi}
\end{figure}

One approach to mitigate this challenge is using partially Bayesian neural networks, treating only a subset of weights probabilistically~\cite{sharma2023partialBnns}.
This significantly reduces training complexity compared to full Bayesian models.
Determining the appropriate number of probabilistic parameters remains an open question, as excessive reduction can degrade predictive performance.
Moreover, partial BNNs typically still require multiple forward passes for uncertainty estimation, limiting efficiency on resource-constrained devices.
In contrast, last-layer methods avoid multiple forward passes by encoding probabilistic samples into multi-headed architectures~\cite{steger2024rlleWorkshop}.
\citeauthor{steger2024rlleWorkshop} demonstrate that a function-space-based repulsive loss effectively enables probabilistic inference even for pretrained models by only training an ensemble of probabilistic output layers~\cite{steger2024rlleWorkshop}.

A widely used approach to enable probabilistic reasoning in neural networks is Stochastic Variational Inference (SVI), which approximates the posterior distribution of weights by assuming a parameterized family of distributions~\cite{murphy2012probabilisticPerspective}, typically Gaussians.  
While SVI provides a tractable framework for Bayesian inference and is more efficient and GPU-friendly compared to Markov Chain Monte Carlo (MCMC) due to its reliance on gradient-based optimization and backpropagation, it still incurs substantial computational costs from weight sampling and the need for multiple forward passes~\cite{jospin2022handsOnBnns}.
An alternative solution is to adopt an analytical approximation, such as the Probabilistic Forward Pass (PFP)~\cite{roth2016pfp}, which represents an extreme case of simplification by assuming not only Gaussian-distributed weights but also Gaussian-distributed activations.
This approximation enables a closed-form solution for uncertainty estimation.  
While still treating all weights in a probabilistic manner, PFP requires only a single, albeit more complex, forward pass for both prediction and uncertainty estimation, eliminating the need for weight sampling and multiple forward passes.

Uncertainty in probabilistic machine learning is categorized as either \emph{aleatoric} or \emph{epistemic}~\cite{kendall2017uncertainties}.  
Aleatoric uncertainty originates from inherent noise in the data and is irreducible, while epistemic uncertainty reflects model uncertainty due to limited knowledge and can be reduced with more data or better modeling.  
For classification problems, aleatoric uncertainty is typically quantified using Softmax Entropy (SME), while epistemic uncertainty---including out-of-domain samples---is measured using Mutual Information (MI).
We refer to Section~\ref{sec:bg:uncertainty_types} for a detailed discussion.  

Figure~\ref{fig:example_samples} illustrates the performance of a SVI-based BNN on images from three representative datasets.
The training dataset MNIST~\cite{Lecun1998mnist} represents in-domain data, while Ambiguous-MNIST~\cite{Mukhoti2022dirtyMNIST} introduces aleatoric uncertainty through samples between classes, and Fashion-MNIST~\cite{xiao2017fashionMnist} contains OOD images of fashion items.
In the following, we will refer to these datasets in combination as "Dirty-MNIST".
Aleatoric uncertainty, quantified via SME, appears for ambiguous samples seen within one sample between classes; OOD images trigger confident but mutually disagreeing predictions.
For each SVI sample, the raw unnormalized outputs, called logits, are typically normalized using a softmax function, and the most likely class is predicted by selecting the maximum value.
Disagreement across predictions, quantified by MI, signals epistemic uncertainty, indicating potential OOD data.

Figure~\ref{fig:number_of_samples_vi} demonstrates that reliable uncertainty quantification requires many samples, causing significant computational demands in traditional BNNs.
For illustrative purposes, only three samples are shown in Figure~\ref{fig:example_samples}.  
The Figure also includes a Gaussian approximation that summarizes the logit samples using their mean and standard deviation.
While this approach somewhat obscures detailed aleatoric and epistemic distinctions, it offers significant compactness benefits.
PFP consistently utilizes this strategy through a single distribution-propagating forward pass.
In practice, it effectively predicts and distinguishes both uncertainty types.

However, since PFP uses non-standard probabilistic operators, an efficient hardware-specific implementation of these operators is required before the method can be used in practice.
These custom operators can be implemented using deep learning compilers, such as TVM~\cite{Chen2018tvm}, which enable automatic code generation and performance tuning~\cite{Shao2022tvmMetaScheduler}.
By leveraging these tools, the rather complex operations of PFP can be optimized for specific hardware architectures, allowing BNNs to run more efficiently in resource-limited environments while maintaining robust uncertainty estimation.

In detail, this work makes the following contributions:
\begin{itemize}  
	\item We propose a staged SVI-to-PFP training pipeline combining Stochastic Variational Inference pre-training, uncertainty calibration, and PFP fine-tuning, demonstrating that the Probabilistic Forward Pass can achieve competitive uncertainty disentanglement and strong out-of-domain detection even for complex architectures and datasets.
    \item We extend the deep learning compiler TVM by integrating essential probabilistic operators for multi-layer perceptron (MLP) and convolutional neural network (CNN) architectures, enabling the efficient execution of PFP-based vision models on ARM processors.  
	\item We implement optimized execution schedules as well as manual and automatic tuning strategies for computationally intensive PFP operators, ensuring high performance on resource-constrained ARM CPUs.
    \item We evaluate the practical feasibility of our approach through benchmarks on embedded ARM processors, demonstrating significant performance improvements that facilitate the deployment of BNNs for uncertainty-aware edge and IoT applications.  
	\item Finally, we investigate the combination of PFP with last-layer Bayesian neural networks, providing initial insights into the trade-offs between uncertainty estimation quality and computational efficiency.
\end{itemize}  

As we will see, although the PFP concept relies on highly non-standard operators, casting the problem within a mapping framework such as TVM enables deployment without specialized libraries while also allowing performance optimizations with reasonable effort.  
The presented PFP operator library is publicly available.\footnote{The code is available at: \url{https://github.com/HAWAIILAB/PFP-Operator-Library}}

%% file: 02-related-work.tex
\section{Background and Related Work}
\label{sec:rw}
This section provides a brief introduction to BNN methods, with a particular focus on related work concerning efficient BNN inference.  
We further clarify the distinction between different types of uncertainty and review methods for efficient inference on resource-constrained embedded systems, specifically in the context of BNNs.

\subsection{Bayesian Neural Networks (BNNs)}
Bayesian neural networks integrate Bayesian inference (BI) into neural networks by treating weights as random variables rather than fixed values~\cite{neal1996bayesian}.  
Weights are sampled from distributions, enabling uncertainty estimation and improved generalization, particularly in data-scarce or noisy settings.  
BNNs place a prior distribution over the network's weights, and then, given observed data, compute a posterior distribution.
The prior reflects any previous knowledge about the weight values, while the posterior captures the updated beliefs after seeing the data~\cite{murphy2012probabilisticPerspective,jospin2022handsOnBnns}.
This probabilistic framework quantifies predictive uncertainty, making BNNs valuable for uncertainty-sensitive and safety-critical tasks.  

Computing the posterior distribution of weights is in general analytically intractable due to the high-dimensional and non-convex nature of neural networks~\cite{jospin2022handsOnBnns}.  
Training and prediction thus generally involves sampling from the weight distributions, unlike deterministic networks that use fixed point estimates.  
This enables BNNs to produce a distribution of possible outcomes rather than a single prediction,
facilitating uncertainty estimation and out-of-domain data detection.  

The equivalent of training in this context is performing Bayesian inference to learn this weight distribution~\cite{jospin2022handsOnBnns}.
BI techniques are commonly categorized into two classes: sampling-based methods and variational methods that optimize an approximate posterior distribution.  
Markov Chain Monte Carlo (MCMC) methods, for instance, sample directly from the posterior distribution~\cite{brooks2011handbook}.
These methods construct a Markov chain with the target posterior as its equilibrium distribution and are often considered the gold standard in terms of quality and reliability.
However, MCMC methods suffer from high computational costs due to slow convergence in high-dimensional spaces and the requirement for a large number of sequential samples.

Variational Inference (VI) reformulates BI as an optimization problem, improving scalability for large datasets and models \cite{blei2017variational,hoffman2013svi}.
It introduces a variational distribution $q_\phi(\theta)$ to approximate the posterior $p(\theta \mid \mathcal{D})$ and minimizes the Kullback-Leibler (KL) divergence $\text{KL}(q_\phi(\theta) \parallel p(\theta \mid \mathcal{D}))$.
The optimization maximizes the Evidence Lower Bound (ELBO): $\mathbb{E}_{q_\phi(\theta)}[\log p(\mathcal{D} \mid \theta)] - \text{KL}(q_\phi(\theta) \parallel p(\theta|\mathcal{D}))$,
\\\noindent which balances data fit and prior regularization \cite{kingma2013auto}.
As an extension of VI, Stochastic Variational Inference (SVI) leverages mini-batches and gradient-based optimization, enabling efficient training on GPUs and scalability to deep neural architectures.
Thus, SVI is a popular choice in BNN research allowing for improved scalability and faster convergence in modern deep learning tasks.

\paragraph{Bayesian Neural Networks and Efficient Inference}
Over the years, a wide range of methodologies for BNNs has been developed; for a comprehensive survey, we refer to \cite{jospin2022handsOnBnns, murphy2012probabilisticPerspective}.
In the following, we highlight some of the most pioneering works that focus on approximate inference techniques designed to enable efficient prediction with BNNs.
While SVI and MCMC methods are mathematically grounded many alternative approaches only loosely relate to these foundational principles.
This often enables improved computational efficiency, but comes at the cost of reduced theoretical rigor and degraded uncertainty estimation performance.

To reduce computational cost, \citeauthor{blundell2015} proposed an SVI-based approach that approximates the posterior using a simplified distribution~\cite{blundell2015}.
This approach reduces computational costs compared to MCMC but remains challenging for deployment on resource-constrained platforms such as mobile devices.  
Hernandez-Lobato and Adams proposed Probabilistic Backpropagation (PBP), a scalable approximate Bayesian inference method that models weight posteriors with factorized Gaussian distributions and propagates activation statistics analytically through the network~\cite{lobato2015probabilisticbackpropagation}. 
This probabilistic propagation concept is closely related to the Probabilistic Forward Pass (PFP) formulation discussed later in Section~\ref{sec:pfp:concept}. 
However, PBP primarily focuses on scalable Bayesian training for fully connected regression models and does not address convolutional architectures or efficient deployment on embedded processors.

\citeauthor{gal2016mcdo} advanced BNN efficiency by proposing Monte Carlo Dropout~\cite{gal2016bcnn,gal2016mcdo}, which approximates BNNs by applying dropout at inference.  
While this is computationally simple, multiple forward passes are still required and uncertainty quality is often significantly reduced.

Deep Ensembles (DE) are another prominent example in this category of algorithms. 
They provide uncertainty estimates by training multiple independent neural networks and aggregating their predictions~\cite{lakshminarayanan2017deepensembles}. 
Due to their simplicity and effectiveness they are widely used.
A fundamental limitation shared with MCMC methods is the substantial memory overhead at inference time, as both approaches require maintaining multiple sets of model parameters or posterior samples, making them impractical for deployment on resource-constrained embedded devices.
Repulsive ensembles~\cite{angelo2021repulsiveensembles} extend Deep Ensembles by promoting diversity among ensemble members.  
\citet{steger2024rlleWorkshop} show that a function-space repulsive loss enables probabilistic inference even for pretrained models by training only an ensemble of probabilistic output layers.

Overall MCDO and ensemble methods have all improved efficient Bayesian inference significantly. 
However, as a common theme they all discard the formal theoretical foundations, found in SVI and MCMC methods.


\subsection{Types of Uncertainty}
\label{sec:bg:uncertainty_types}

In the probabilistic ML context uncertainty is categorized into \emph{aleatoric and epistemic uncertainty}~\cite{kendall2017uncertainties}.  
\emph{Aleatoric uncertainty} stems from inherent noise in the data, making it irreducible even with unlimited data.  
It arises when input features contain measurement noise or the output variable is intrinsically unpredictable.  

\emph{Epistemic uncertainty} arises from a lack of knowledge, such as missing data or suboptimal models.  
It is reducible and decreases with more data or model refinement.  
This uncertainty is prominent in underrepresented input regions, including OOD data.  
The key distinction is that aleatoric uncertainty is data-driven and irreducible, while epistemic uncertainty is model-related and can be reduced.  

BNNs distinguish between aleatoric and epistemic uncertainty in both regression and classification tasks.  
In regression, aleatoric uncertainty is modeled via a heteroscedastic head that predicts data-dependent noise.  
For classification, aleatoric and epistemic uncertainty are commonly estimated via Softmax Entropy and Mutual Information, respectively~\cite{Mukhoti2022dirtyMNIST}.

As shown in Figure~\ref{fig:example_samples}, each predictive sample in classification yields distinct predicted logits.  
For classification, total predictive uncertainty is quantified using \emph{Shannon Entropy}~\cite{shannon1948entropy}:
\begin{align}
	\mathbb{H}(y|x,\mathcal{D}) &= - \sum_c^K \left[ \frac{1}{N}\sum_n^N \left( p(y=c|x,w_n) \right) \cdot \text{log}\left[ \frac{1}{N}\sum_n^N \left( p(y=c|x,w_n) \right) \right] \right]
\end{align} 

Here $p(y=c|x,w_n)$ denotes the probability of prediction $y$ and ground-truth class $c$ being identical, given input $x$ and weight sample $w_n$ from the $n$-th sample. 
$\mathcal{D}$ denotes the full dataset, $K$ the number of classes, and $N$ the number of samples.
This total predictive uncertainty is decomposed into their aleatoric and epistemic components, measured by the \emph{Softmax Entropy} for aleatoric uncertainty,
\begin{equation}
	\mathbb{E}_{p(w|\mathcal{D})}\big( \mathbb{H}(y|x,w) \big) = - \frac{1}{N} \sum_n^N \sum_c^K p(y=c|x,w_n) \cdot \text{log}\big( p(y=c|x,w_n) \big)
\end{equation}
and \emph{Mutual Information}~\cite{Depeweg2018} for epistemic uncertainty
\begin{equation}
	\mathbb{I}(y,w|x,\mathcal{D}) = \mathbb{H}(y|x,\mathcal{D}) - \mathbb{E}_{p(w|\mathcal{D})}\big( \mathbb{H}(y|x,w) \big).
\end{equation} 


\subsection{Efficient Inference of Neural Architectures on Resource-constrained Devices}

Deploying modern machine learning models on mobile processors requires careful optimization for computational and memory constraints.  
To this end \citeauthor{Chen2018tvm} introduced TVM~\cite{Chen2018tvm}, an automated deep learning compiler that optimizes models for diverse hardware backends, including mobile ARM processors.  
TVM facilitates efficient deployment of deep neural architectures by automating operator specific optimizations and supporting hardware-specific enhancements.

Quantization~\cite{Jacob_2018_CVPR} reduces the precision of neural network weights and activations, significantly lowering inference resource requirements.  
Deep compression methods, including pruning and quantization~\cite{han2016deepcompression, schindler2018}, reduce the resource footprint, enhancing applicability to mobile processors.
Neural architecture search and automatic compression, combined with reinforcement learning~\cite{caiProxylessNASDirectNeural2019, caiZeroQNovelZero2020, heAMCAutoMLModel2019, krieger2022}, automate architecture refinement and compression parameter selection, optimizing bit width and sparsity, and enabling compression at the layer, channel, or block level.  
For a more detailed overview of resource-efficient neural network inference on embedded systems, including a comparison of various compression methods and hardware architectures, we refer to \cite{roth2024rennfes}.

Considering the intersection of Bayesian methods and neural networks from a different perspective, \citeauthor{louizos2017} propose Bayesian compression techniques for DNNs~\cite{louizos2017}, leveraging priors to guide pruning while using posterior uncertainties to determine optimal fixed-point precision.  
Although relevant for mobile DNN deployments, this approach does not address the efficiency of BNNs.  

\paragraph{Bayesian Neural Networks on Resource-Constrained Devices}
Although direct research on deploying BNNs on mobile processors is limited, several works address related areas.  
A closely related effort is by \citeauthor{banerjee2019}, who propose AcMC2, a compiler that transforms probabilistic models into optimized MCMC-based hardware accelerators, such as FPGAs or ASICs~\cite{banerjee2019}.  
However, this work targets general probabilistic models, which differ significantly in scale and underlying principles from BNNs.

For specialized accelerators based on standard CMOS technology, ShiftBNN~\cite{shiftbnn}, $B^2N^2$~\cite{awano2023b2n2} and VIBNN~\cite{cai2018} leverage ASICs and FPGAs, respectively, to accelerate BNN training and inference using large-scale Gaussian random number generators.  
Beyond CMOS, early research explores alternative technologies, such as analog computing and resistive memory, to introduce stochasticity as a source of uncertainty in Bayesian methods.  
Examples include BNNs on probabilistic photonic hardware~\cite{brckerhoffplckelmann2024probabilistic}, custom hardware for probabilistic circuits~\cite{yao2024HwPc}, and Bayesian networks utilizing resistive memory and its inherent operational uncertainty~\cite{8993616, Bonnet2023, 9773213}.  
However, these studies remain in early stages with limited experimental validation, leaving the feasibility of these technologies uncertain.  

In summary, the intersection of BNNs and mobile processors remains an emerging area of research. 
While foundational methods such as MCDO and SVI offer pathways toward efficient uncertainty estimation, the challenge of deploying these methods on resource-constrained devices like mobile ARM processors is still largely unexplored. 
Tools like TVM, combined with techniques such as quantization and model partitioning, provide promising avenues for bringing the benefits of Bayesian inference to mobile platforms. 
Our work aims to build on these foundations by mapping efficient Stochastic Variational Inference methods to mobile ARM processors using TVM, thereby addressing the unique challenges posed by this constrained environment.

%% file: 03-pfp-concept.tex
\section{Probabilistic Forward Pass}
\label{sec:pfp:concept}

\begin{wrapfigure}{r}{0.3\textwidth}
    \centering
	\vspace{-0.6cm}
    \includegraphics[width=\linewidth]{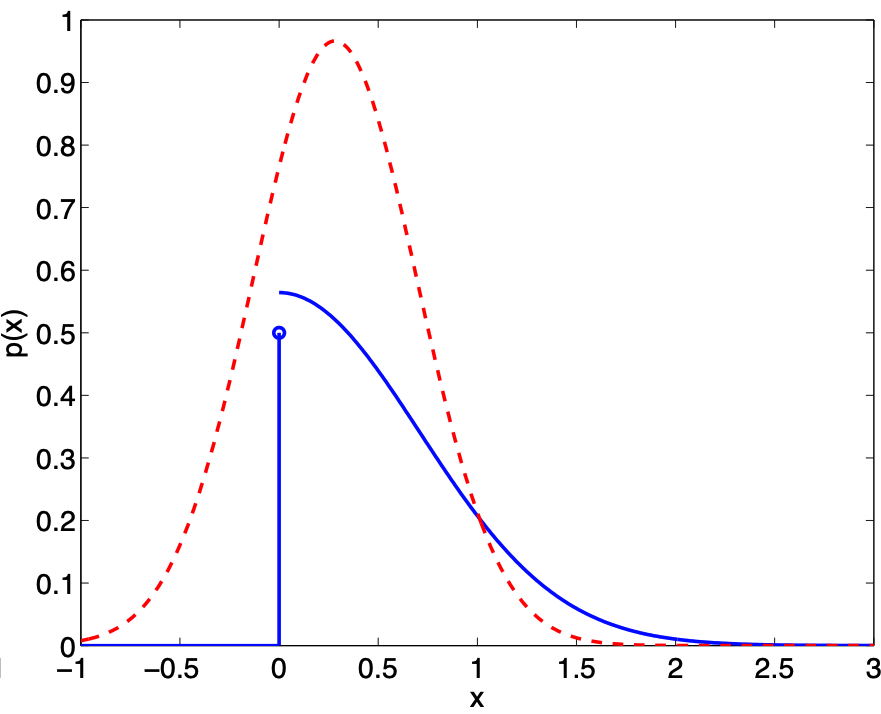}
    \caption{Illustration of Gaussian moment matching for ReLU activations. The true distribution (solid line) is approximated as a Gaussian (dashed line). Reproduced with permission from~\cite{roth2016pfp}.}
    \label{fig:relu_moment_matching}
	\vspace{-0.5cm}
\end{wrapfigure}
The Probabilistic Forward Pass~\cite{roth2016pfp,roth2021phd} efficiently implements BNNs by propagating probability distributions through the network, eliminating the need for repeated sampling and multiple forward passes.  
This approach significantly reduces computational overhead while preserving predictive uncertainty estimation.  

PFP represents an extreme approximation of Stochastic Variational Inference-based BNNs (SVI-BNNs).  
While SVI-BNNs typically assume Gaussian-distributed weights with a mean-field independence assumption, PFP extends this to Gaussian-distributed activations.  
This assumption leverages the central limit theorem, which states that the sum of many independent random variables tends toward a normal distribution.  
Unlike general SVI, which can model complex activation distributions, PFP constrains activations to a Gaussian form.  
This restriction enables a single closed-form forward pass, eliminating the need for stochastic sampling and multiple evaluations.  
As a result, PFP is particularly suited for resource-constrained environments, where computational efficiency and reliable uncertainty estimation are critical.  
Input distributions are transformed into output distributions through linear and non-linear operations adapted to handle Gaussian distributions.  
Linear layers, batch normalization, and residual skip connections can be handled through affine or additive moment propagation, whereas non-linear activations require additional approximations.

Consider a fully connected layer where the input follows a Gaussian distribution with mean and variance \((\mu_{\text{in}}, \sigma_{\text{in}}^2)\), and the weights are Gaussian with parameters \((\mu_w, \sigma_w^2)\).  
The output pre-activation distribution is derived by computing the mean and variance of the linear transformation.  
The mean is obtained by propagating expectations, while the variance incorporates uncertainties from both weights and inputs.  
PFP considers activations to be independent, ensuring variance computations remain tractable.  
Equations~\ref{eq:pfp_dense_scalar:mean} and~\ref{eq:pfp_dense_scalar:variance} illustrate the scalar computation of mean and variance for the $l$-th fully connected layer,  

\begin{align}
  \mu_{a_i^l} &= \sum_{j=1}^{d_{l-1}} \mu_{w^l_{ij}} \cdot \mu_{x_j^{l-1}} \label{eq:pfp_dense_scalar:mean} \\
	\sigma_{a_i^l}^2 &= \sum_{j=1}^{d_{l-1}} \sigma_{w_{ij}^l}^2 \cdot \mathbb{E} \left[ \left(x^{l-1}_j \right)^2 \right] + \mu_{w_{ij}^l}^2 \cdot \left( \mathbb{E} \left[ \left(x^{l-1}_j \right)^2 \right] - \mu_{x_j^{l-1}}^2 \right). \label{eq:pfp_dense_scalar:variance}
\end{align}
Thereby the \emph{second raw moment} $\mathbb{E}(x^2)=\mu^2 + \sigma^2$ is used.
Moreover, $d_{l-1}$ denotes the width of the previous layer, which is the input activation tensor width of the current layer.
Reformulation to use means and variances instead of \emph{second raw moments} is possible,
\begin{align}
  \bm{\sigma_a^2} &= \bm{\sigma_w}^2 \cdot \mathbb{E}[\bm{x}^2] + \bm{\mu_w}^2 \cdot \big( \mathbb{E}[\bm{x}^2] - \bm{\mu_x}^2 \big) \label{eq:pfp_dense_vectorized:mE} \\
                  &= \bm{\sigma_w}^2 \cdot \bm{\mu_x}^2 + \bm{\mu_w}^2 \cdot \bm{\sigma_x}^2 + \bm{\sigma_w}^2 \cdot \bm{\sigma_x}^2 \label{eq:pfp_dense_vectorized:mv}
\end{align}
illustrated here in tensor notation.
Later implementations support both variants, selecting the one with lower computational overhead.  
To avoid unnecessary conversions, the output of one layer and the input of the next must be consistently represented, either as mean and variance or as mean and \emph{second raw moment} tuples.  

A key challenge in propagating distributions arises from non-linear activation functions.  
For the widely used Rectified Linear Unit (ReLU), PFP approximates post-activation outputs by matching the first two moments (mean and variance) of the original output distribution.  
Figure~\ref{fig:relu_moment_matching} illustrates this moment-matching process, where an input truncated Gaussian is transformed back into a proper Gaussian distribution.  
Thus, the PFP-specific ReLU operator, while still an element-wise operation, is more complex than its deterministic counterpart, $ReLU(x) = \max(0, x)$,  
\begin{align}
  \mu_{x_i^l} &=  \mathbb{E}[x_i^l] = \frac{\mu_{a_i^l}}{2} \left( 1 + \text{erf} \left( \frac{\mu_{a_i^l}}{\sqrt{2 \sigma_{a_i^l}^2}} \right) \right) + \sqrt{\frac{\sigma_{a_i^l}^2}{2 \pi}} \exp \left( - \frac{\mu_{a_i^l}^2}{2 \sigma_{a_i^l}^2} \right) \\
  \mathbb{E} \left[ (x_i^l)^2 \right] &= \frac{\sigma_{a_i^l}^2 + \mu_{a_i^l}^2}{2} \left(  1 + \text{erf} \left( \frac{\mu_{a_i^l}}{\sqrt{2 \sigma_{a_i^l}^2}} \right) \right) + \mu_{a_i^l} \sqrt{\frac{\sigma_{a_i^l}^2}{2 \pi}} \exp \left( - \frac{\mu_{a_i^l}^2}{2 \sigma_{a_i^l}^2} \right) 
  \label{eq:pfp:relu}
\end{align}
where erf denotes the error function $\text{erf(u)} = \frac{2}{\sqrt{\pi}}\int_0^u \exp (- z^2 ) dz$~\cite{roth2021phd}.
As distributions propagate through successive layers, the output layer generates the final predictive distribution.  

A key advantage of PFP is its ability to compute the expected log-likelihood in closed form, eliminating the need for sampling.  
At test time, predictions require only a single forward pass, directly computing expected outputs and uncertainties without costly sampling or model averaging.  
By reformulating neural network computations to operate on distributions, PFP offers a scalable and efficient approach for deploying BNNs in practical settings.  

\subsection{Conceptual Limitations}
\label{subsec:limitations}

PFP approximates Bayesian inference by analytically propagating Gaussian-distributed activations through the network.
This enables highly efficient single-pass inference, but restricts the expressiveness of the resulting predictive distributions compared to sampling-based approaches such as SVI.
In particular, epistemic uncertainty estimation can degrade when the true activation or logit distributions deviate substantially from a Gaussian form.
Such deviations may become increasingly relevant for larger architectures and more complex datasets, where predictive distributions can exhibit highly non-linear or multi-modal behavior.

Consequently, while PFP provides uncertainty estimates closely aligned with SVI for many practical scenarios, the Gaussian activation approximation introduces an inherent trade-off between computational efficiency and uncertainty representation fidelity.
As discussed later in Section~\ref{sec:pfp:evaluation}, these limitations become more apparent for increasingly challenging classification tasks.

%% file: 04-pfp-training.tex
\section{PFP Training and Uncertainty Estimation Experiments}
\label{sec:pfp:training}
A key advantage of PFP is its compatibility with pretrained SVI-based Bayesian neural networks.
This enables the proposed \emph{staged SVI-to-PFP training pipeline}, which combines the stable optimization behavior of SVI with the efficient inference capabilities of PFP.
Probabilistic Programming Languages (PPLs) such as Pyro~\cite{bingham2019pyro} and TensorFlow Probability\footnote{\url{https://github.com/tensorflow/probability}} provide efficient frameworks for training probabilistic neural networks using SVI.
In this work, all SVI-BNNs are trained using Pyro and subsequently transferred to the corresponding PFP formulation.

\subsection{Staged SVI-to-PFP Training Pipeline}
Direct training of PFP-based BNNs, as originally proposed in~\cite{roth2016pfp}, was found to be highly sensitive to hyperparameters and optimization settings in our experiments.
While direct PFP training remains feasible for smaller architectures and simpler datasets, Table~\ref{tab:auroc} shows that its performance deteriorates substantially for more complex architectures and datasets such as ResNet-18 on CIFAR-10 and CINIC-10.
Pure PFP training was therefore not sufficiently robust for the larger-scale settings considered here.

We therefore propose a novel training pipeline.
The new pipeline is conducted in stages and first trains a BNN using SVI, yielding Gaussian posterior distributions over all probabilistic weights.
The learned posterior means and variances are subsequently transferred into the corresponding PFP network, in theory only requiring a conversion from logarithmic to standard variance representation.

However, directly transferring the learned SVI posterior to PFP generally results in suboptimal uncertainty estimates.
To compensate for the mismatch between stochastic SVI inference and the analytical PFP approximation, all transferred variances are globally rescaled using a heuristic calibration based on OOD-detection performance.
The resulting calibrated PFP model is reported separately in Table~\ref{tab:auroc}.

In a final stage, the calibrated PFP model is fine-tuned directly using the PFP objective.
During this stage, the relative weighting of the KL regularization term and the variance regularization parameter are optimized using hyperparameter optimization.
Due to the multi-objective nature of PFP training, longer optimization does not necessarily improve predictive accuracy and uncertainty estimation quality simultaneously.
In practice, extended fine-tuning frequently improved one objective while degrading the other.
Consequently, the fine-tuning stage is monitored throughout training and terminated once uncertainty estimation quality saturates.
As shown in Table~\ref{tab:auroc}, the new staged SVI-to-PFP pipeline exhibits substantial quality improvements over direct PFP-only training, while also being more stable.
This is particularly significant for larger architectures and more complex datasets.

\subsection{Experimental Setup}
Three neural architectures were used in the experiments:
a simple multi-layer perceptron (MLP) with one hidden layer of 100 neurons, the LeNet-5~\cite{Lecun1998mnist} architecture, and the ResNet-18~\cite{ResNet} architecture.
Weights are treated probabilistically using Gaussian priors and posteriors with learnable parameters.
Additionally, the mean-field assumption~\cite{farquhar2020meanfield} is employed, assuming independent Gaussian weight distributions.

To evaluate the BNNs' ability to distinguish between aleatoric and epistemic uncertainty, we use the Dirty-MNIST dataset introduced by \citeauthor{Mukhoti2022dirtyMNIST}, which extends MNIST with Ambiguous-MNIST for aleatoric uncertainty and Fashion-MNIST for epistemic uncertainty.
Following their experimental protocol, training is performed on MNIST together with Ambiguous-MNIST, while Fashion-MNIST remains entirely unseen during training~\cite{Mukhoti2022dirtyMNIST}.

To evaluate scalability to more complex datasets and architectures, we use CIFAR-10~\cite{CIFAR10} and CINIC-10~\cite{CINIC10} as in-domain datasets and Street View House Numbers (SVHN)~\cite{SVHN} for out-of-domain evaluation.
Training SVI-BNNs on ImageNet~\cite{IMAGENET} remains prohibitively expensive (1.2M images and 1{,}000 classes versus 60k images and 10 classes for CIFAR-10), which also limits the scalability of our PFP pipeline, as it relies on SVI-trained models.
We therefore use CINIC-10 as an intermediate step between CIFAR-10 and ImageNet.
Although its image resolution matches that of CIFAR-10, CINIC-10 contains downscaled images from ImageNet, making the classification task significantly more challenging.
For these scaling evaluations, we use a BNN based on the ResNet-18 architecture~\cite{ResNet}.

\subsection{SVI Training and ResNet-18 Scaling}
SVI training closely resembles standard neural network optimization using gradient descent, but introduces additional complexity through the probabilistic objective formulation.
Due to the multi-objective nature of the ELBO optimization, training is generally more sensitive to initialization and hyperparameter selection than deterministic training.

The MLP and LeNet-5 SVI-BNNs are trained for $1000$ epochs using the Adam optimizer~\cite{kingma2014adam} with a constant learning rate of $0.001$.
The variational weights are initialized with $\mu=0.08$ and $\sigma=0.0001$, and a mini-batch size of $100$ is used.

Balancing the expected log-likelihood term and the KL-divergence term in the ELBO using a constant factor $\alpha$ proved challenging.
We therefore employ \emph{KL annealing}~\cite{abrol2014deterministic,Zhang2019advancesVI}, where the KL contribution is increased gradually throughout training.
The dynamically adapted ELBO,
\begin{equation}
	\text{ELBO}(q,e) =
	\mathbb{E}[\log p(\mathcal{D}|w)]
	-
	A(e)\cdot \text{KL}(q(w)||p(w|\mathcal{D})),
\end{equation}
uses a linearly increasing KL factor $A(e)$ that scales from $0$ to $\alpha_{\text{max}}=0.25$ over the training epochs $e$.
KL annealing substantially improves optimization stability and reduces sensitivity to initialization.

Training probabilistic ResNet-18 architectures on CIFAR-10 and CINIC-10 introduces substantially higher optimization complexity compared to the MLP and LeNet-5 experiments.
SVI-BNN training for larger architectures is computationally expensive and adds numerous additional sensitive hyperparameters.
To make these experiments feasible within acceptable training times, the training pipeline was extended substantially.

First, a deterministic ResNet-18 is trained on the respective dataset for $400$ epochs using the schedule-free version~\cite{schedulefree} of the RAdam optimizer~\cite{RADAM} with a learning rate of $0.00215$.
The resulting deterministic weights are then loaded into the SVI-BNN model as the initial mean values of the Gaussian variational posterior.

As demonstrated in previous work~\cite{sharma2023partialBnns}, the first and last layers of BNNs contribute disproportionately to predictive uncertainty quality.
We therefore assign different prior standard deviations to the first and last layers of the ResNet-18 architecture, while intermediate layers share a common prior configuration.

Due to the large number of sensitive hyperparameters, hyperparameter optimization (HPO) is performed using the Multi-Objective Tree-structured Parzen Estimator~\cite{TPE}, as implemented in Optuna~\cite{optuna}.
For CIFAR-10, the HPO is executed in two stages comprising 56 and 54 trials, respectively, gradually refining the search ranges.
Subsequently, an additional HPO consisting of 42 trials is conducted for CINIC-10 using refined search ranges and pretrained deterministic initialization.
For Dirty-MNIST, deterministic pretraining was omitted due to the reduced dataset complexity, while the ResNet-18 HPO pipeline was otherwise largely reused with an increased number of HPO trials (130).
The resulting optimal hyperparameter configurations for all evaluated datasets are provided in the accompanying code repository\footnote{\url{https://github.com/HAWAIILAB/PFP-Operator-Library/blob/main/hyperparameter_optimization/SVI_ResNet.md}}.

\subsection{Qualitative Performance}
\label{sec:pfp:evaluation}

\begin{DIFnomarkup}
\begin{table}
    \centering
    \caption{Classification accuracy and OOD-detection AUROC for SVI and PFP-based training variants.}
    \label{tab:auroc}
    \small
	\begin{diffbox}
	\begin{tabular}{l|l| l| c| c|c}
        \hline
        \textbf{Architecture} & \textbf{Dataset} & \textbf{Method} & \textbf{Calib.} & \textbf{Acc. (\%)} & \textbf{AUROC (\%)} \\
        \hline

        MLP & Dirty-MNIST & SVI & -- & 96.30 & 81.2 \\
			&             & PFP-only training (100 epochs) & - & 97.71 & 85.4 \\ 
            &             & SVI $\rightarrow$ PFP, calibrated & 0.3 & 96.30 & 85.8 \\
			&             & SVI $\rightarrow$ PFP, calibrated + fine-tuned (6 epochs) & 0.3 & 97.08 & 90.4 \\ 
        \hline

        LeNet-5 & Dirty-MNIST & SVI & -- & 98.70 & 98.6 \\
				&             & PFP-only training (200 epochs) & -- & 97.64 & 80.5 \\ 
                &             & SVI $\rightarrow$ PFP, calibrated & 0.4 & 98.90 & 96.6 \\
				&             & SVI $\rightarrow$ PFP, calibrated + fine-tuned (50 epochs) & 0.4 & 98.54 & 98.5 \\
        \hline

        ResNet-18 & Dirty-MNIST & SVI & -- & 99.37 & 99.4 \\
				  &             & PFP-only training (100 epochs) & -- & 98.72 & 85.1 \\ 
                  &             & SVI $\rightarrow$ PFP, calibrated & 0.6 & 99.22 & 94.0 \\ 
				  &             & SVI $\rightarrow$ PFP, calibrated + fine-tuned (1 epoch) & 0.6 & 99.30 & 94.0 \\ 
        \hline

        ResNet-18 & CIFAR-10 & SVI & -- & 91.52 & 89.2 \\
				  &          & PFP-only training (200 epochs) & -- & 73.72 & 73.0 \\ 
				  &          & SVI $\rightarrow$ PFP, calibrated & 0.4 & 91.50 & 94.1 \\ 
				  &          & SVI $\rightarrow$ PFP, calibrated + fine-tuned (10 epochs) & 0.4 & 91.20 & 95.9 \\ 
        \hline

        ResNet-18 & CINIC-10 & SVI & -- & 81.49 & 86.2 \\
				  &          & PFP-only training (200 epochs) & -- & 32.29 & 51.5 \\ 
                  &          & SVI $\rightarrow$ PFP, calibrated & 0.45 & 62.65 & 75.7 \\
				  &          & SVI $\rightarrow$ PFP, calibrated + fine-tuned (10 epochs) & 0.45 & 63.86 & 78.3 

    \end{tabular}
	\end{diffbox}

\end{table}
\end{DIFnomarkup}

\begin{figure}  
    \centering  
    \includegraphics[width=0.95\textwidth]{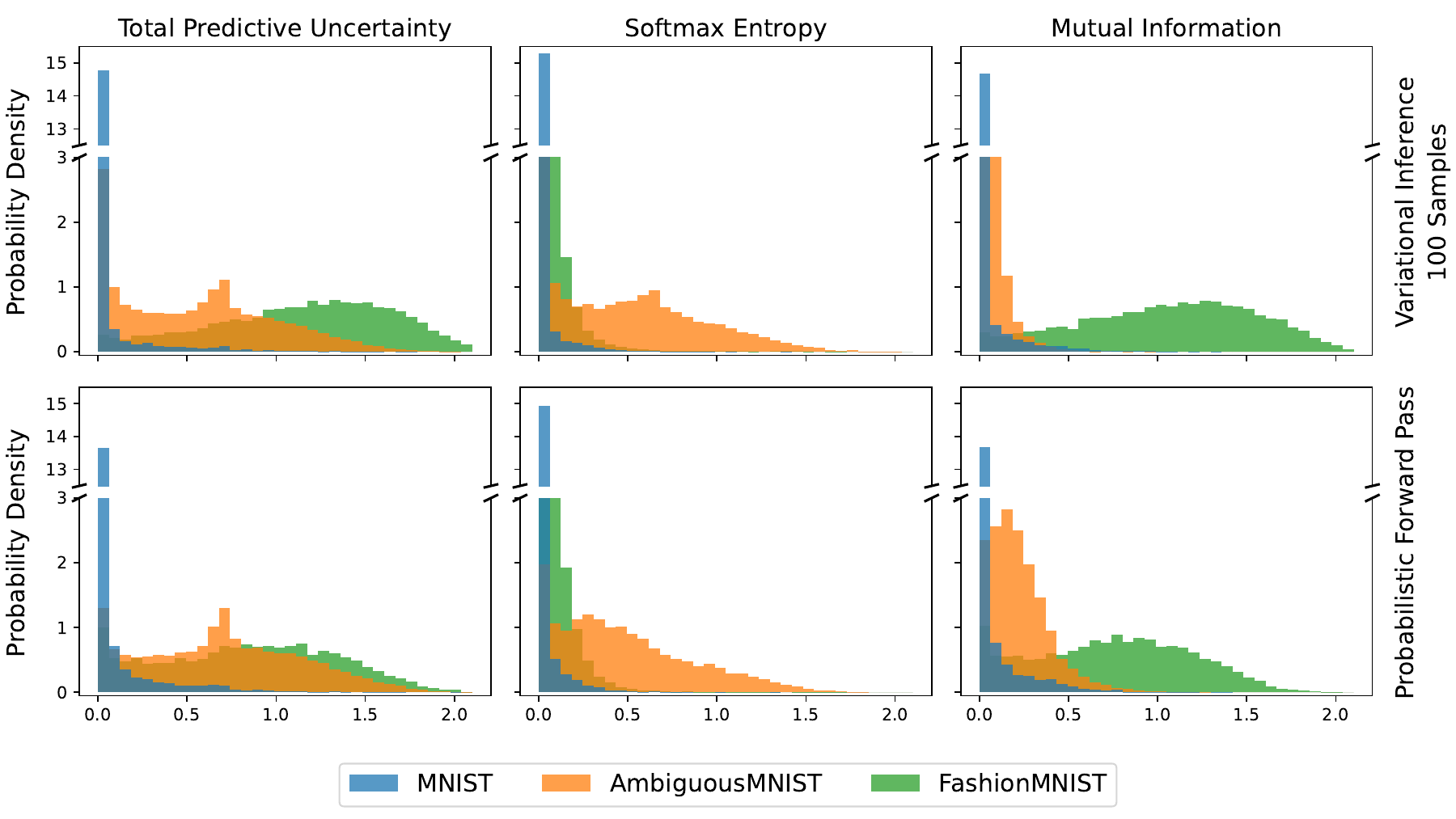}  
    \caption{Comparison of SVI and PFP uncertainty predictions.  
    For MNIST, both uncertainties are expected to be low; Ambiguous-MNIST exhibits higher aleatoric uncertainty (Softmax Entropy), and Fashion-MNIST, as OOD data, shows higher epistemic uncertainty (Mutual Information).  
    Both methods effectively assign the majority of images to their respective domains.}  
    \label{fig:hist_vi_pfp_TuSeMi}  
\end{figure}

To evaluate the effectiveness of the proposed staged SVI-to-PFP training pipeline, we compare the resulting PFP-BNNs against their original SVI counterparts on uncertainty estimation and OOD detection tasks.
Figure~\ref{fig:example_samples} illustrates the performance on individual inputs, while Figure~\ref{fig:hist_vi_pfp_TuSeMi} compares key uncertainty metrics: Shannon Entropy for total uncertainty, Softmax Entropy for aleatoric uncertainty, and Mutual Information for epistemic uncertainty.  
These metrics aggregate over the sample dimension, and as shown in Figure~\ref{fig:number_of_samples_vi}, the number of samples significantly impacts Shannon Entropy, thereby influencing Mutual Information.  
Unlike SVI-BNNs, PFP does not inherently provide a sample dimension.  
To ensure a fair comparison, we introduce an artificial sampling dimension using PFP-predicted means (\( \mu_{\text{PFP}} \)) and variances (\( \sigma^2_{\text{PFP}} \)) of the logits.  
Assuming a Gaussian distribution, logit samples $l_{\text{PFP}}$ are generated as $l_{\text{PFP}} \sim \mathcal{N}(\mu_{\text{PFP}}, \sigma^2_{\text{PFP}})$.  
This \emph{logit sampling} approach is computationally efficient, as it avoids sampling within the network and eliminates the need for multiple forward passes.  
It serves as a post-processing step, enabling standard uncertainty metrics typically used for sampling-based BNNs.  
In resource-constrained applications, directly leveraging PFP-predicted variances (\( \sigma^2_{\text{PFP}} \)) for decision-making is an efficient alternative.  

Figure~\ref{fig:hist_vi_pfp_TuSeMi} shows that both methods yield higher total predictive uncertainty for Ambiguous-MNIST and Fashion-MNIST compared to the lower uncertainties observed for standard MNIST.  
Similarly, both exhibit elevated Softmax Entropy for Ambiguous-MNIST, indicating increased aleatoric uncertainty, and higher Mutual Information for the out-of-distribution Fashion-MNIST dataset, reflecting greater epistemic uncertainty.  
Overall, both approaches provide uncertainty estimates consistent with expectations and desired behavior.  

\begin{figure}
    \centering  
    \includegraphics[width=0.6\textwidth]{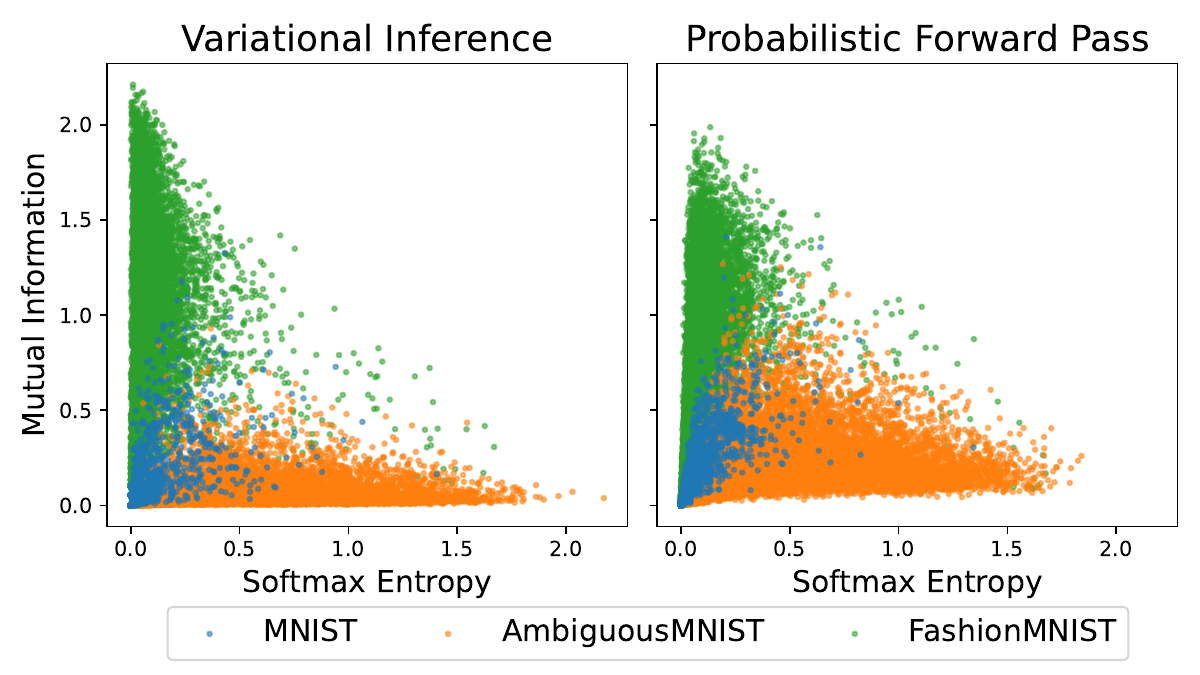}  
    \caption{Disentangled Epistemic (MI) and Aleatoric (SME) Uncertainty.  
    Comparison of SVI and PFP uncertainty predictions.  
    While PFP performs comparably to SVI in estimating total uncertainty, its ability to disentangle aleatoric and epistemic uncertainties is somewhat limited, as anticipated.  
    Nevertheless, the practical distinction remains sufficiently robust for most cases.}  
    \label{fig:scatter_vi_pfp_SeMi}  
\end{figure}

A detailed analysis of Softmax Entropy and Mutual Information across all images, shown in Figure~\ref{fig:scatter_vi_pfp_SeMi}, indicates that SVI outperforms PFP in disentangling aleatoric and epistemic uncertainty, as theoretically expected.  
However, from a practical standpoint, PFP still provides sufficient separation in most cases.  
Only in certain edge cases do both uncertainties exhibit high values, resulting in less distinct separation.  

The progressive improvements observed in Table~\ref{tab:auroc} are also reflected qualitatively in Figures~\ref{fig:hist_vi_pfp_TuSeMi} and~\ref{fig:scatter_vi_pfp_SeMi}, where the calibrated PFP models recover uncertainty distributions closely aligned with those of the original SVI-BNNs.  

The Area Under the Receiver Operating Characteristic Curve (AUROC) is used to evaluate out-of-domain detection performance.
An AUROC of $1.0$ indicates perfect discrimination, while $0.5$ corresponds to random guessing~\cite{fawcett2006roc}.

Table~\ref{tab:auroc} presents a comparative analysis of SVI and PFP using AUROC as a metric to evaluate their effectiveness in OOD detection.  
The results indicate that both methods exhibit comparable performance in terms of predictive accuracy and OOD detection capability.  
However, the stronger performance degradation observed for the more challenging CINIC-10 benchmark is consistent with the conceptual limitations of PFP, suggesting that the Gaussian activation approximation may become insufficient for increasingly complex predictive distributions.

Furthermore, the influence of the neural architecture is evident, as both methods demonstrate improved predictive performance and OOD detection when utilizing a more expressive convolutional neural network.
In summary, PFP achieves prediction quality closely aligned with SVI in both accuracy and OOD detection.

%% file: 05-operator-library.tex
\section{Extending TVM with a PFP Operator Library}
\label{sec:operator_library}
TVM provides multiple internal languages and intermediate representations, including TensorIR~\cite{feng2023tensorir}, TensorExpression (TE), TVMScript, and Relax~\cite{Lai2023relax}, which are essential for implementing custom operators.  
While TE defines computational rules in a very succinct way, TensorIR defines the interaction of computations more fine-grained as modular \emph{blocks}, enabling flexible scheduling and optimizations.\footnote{\url{https://tvm.apache.org/docs/deep_dive/tensor_ir/tutorials/tir_creation.html}}  
Relax~\cite{Lai2023relax}, the successor to Relay~\cite{Roesch2018relay}, serves as a high-level intermediate representation, supporting dynamic shapes, control flow, and seamless integration with TensorIR.\footnote{\url{https://tvm.apache.org/docs/deep_dive/relax/learning.html}} 
TVMScript, a Python-based frontend, allows the direct definition and modification of TensorIR and Relax. 

To implement a custom operator, developers define its computation in TE and create IRModules via the BlockBuilder-API.
BlockBuilder creates primitive functions from TE expressions that are connected via Relax and can be optimized using TensorIR scheduling.\footnote{\url{https://mlc.ai/docs/get_started/tutorials/quick_start.html}}  
This workflow facilitates efficient execution of specialized operators across diverse hardware platforms while minimizing implementation complexity.

\subsubsection*{Operating on Tuples}
Operators with multiple input tensors are common in neural networks, whereas those producing multiple output tensors are relatively rare.  
Typically, operators perform a single core computation uniformly across all data.  
However, PFP introduces a unique requirement, as it maintains separate compute paths and outputs for mean and variance.  
TVM follows the \emph{one operator = one compute rule} principle, ensuring that each operator executes a single, sequential stream of instructions without divergence.  
Consequently, logical PFP operations may be split into separate TVM operators, e.g., one for means and another for variances.  
However, this approach increases the complexity of interconnecting sub-operators, adds overhead, and complicates the implementation of new network architectures.  
Moreover, from a resource efficiency perspective, this separation is undesirable, as mean and variance calculations share common sub-terms.  
Achieving this integration requires extensions to TVM's basic operations such as summation over tuples, however, it enables data reuse and eliminates redundant computations.  
Figure~\ref{fig:lib:operator_implementations} demonstrates that network architectures employing joint operators, as opposed to separate computational paths for mean and variance, consistently benefit from enhanced data reuse.
This structural advantage translates into superior performance across all evaluated cases, indicating the efficacy of the joint formulation in leveraging shared intermediate representations.

\subsubsection*{Variance and Second Raw Moment}
\begin{figure}
    \centering
    \includegraphics[width=0.5\textwidth]{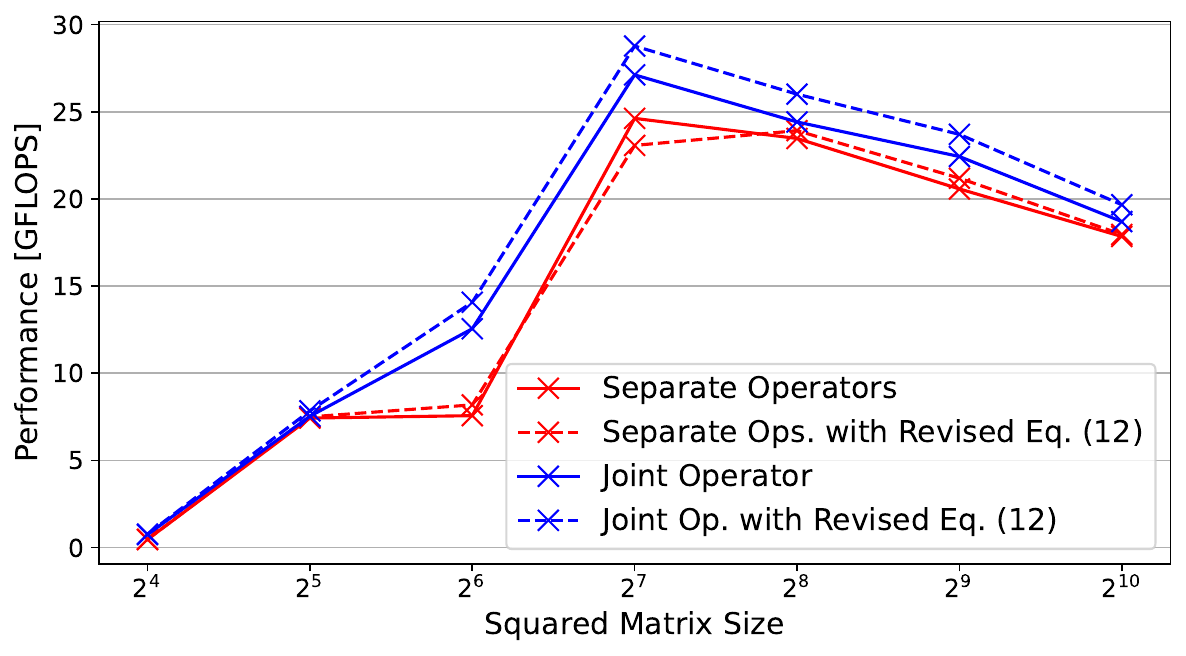}    
	\caption{Performance comparison of operator implementations, evaluating the reformulation from Equation~\ref{eq:pfp_dense_scalar:variance} to \ref{eq:pfp_dense_scalar:E:variance} and the use of separate vs. joint operators for mean and variance paths on a ARM Cortex-A72.}  
    \label{fig:lib:operator_implementations}
\end{figure}

The original formulation of PFP operators is based on mean and variance, with inputs and outputs defined accordingly.  
However, reformulating Equation~\ref{eq:pfp_dense_scalar:variance} to operate on second raw moments for activations and weights enhances data reuse and reduces computational overhead.  
Using second raw moments, the variance in the PFP dense layer can be computed as:  
\begin{equation}
	\sigma_{a_i^l}^2 = \sum_{j=1}^{d_{l-1}} \mathbb{E} \left[ \left(w_{ij}^l\right)^2 \right] \cdot \mathbb{E} \left[ \left( x_{j}^{l-1} \right)^2 \right] - \left( \mu_{w_{ij}^l} \cdot \mu_{x_j^{l-1}}\right)^2, \label{eq:pfp_dense_scalar:E:variance}
\end{equation} thereby reusing the means for the current layer $\mu_{a_i^l}$, the pre-computed second raw moments of the weights $\mathbb{E} [ (w_{ij}^l)^2 ]$ and avoiding conversions from previous activation function outputs $\mathbb{E} [ ( x_{j}^{l-1} )^2 ]$, which compute second raw moments by design.
The locality of this tuple-based second raw moment operator improves cache efficiency and overall performance.  
Figure~\ref{fig:lib:operator_implementations} illustrates the performance gains achieved by reformulating Equation~\ref{eq:pfp_dense_scalar:variance} to \ref{eq:pfp_dense_scalar:E:variance} and employing a joint operator.  

When consecutive layers differ in their representation of second raw moments and variances, conversion is straightforward using $\mathbb{E}(x^2) = \mu^2 + \sigma^2$.  
However, converting second raw moments to variances and then back in subsequent layers introduces unnecessary computational overhead.  
To mitigate this, the operator implementation includes a conversion function as a configurable argument.  
Ensuring consistency between the inputs and outputs of connected layers remains the responsibility of the model designer.  
Moreover, the weights need to be expressed accordingly, either as means and variances $\sigma_w^2$ (see~\eqref{eq:pfp_dense_vectorized:mv}) or as means and second raw moments $\mathbb{E}[w^2]$ (see~\eqref{eq:pfp_dense_scalar:variance}).

To minimize redundant computations and facilitate the reuse of intermediate statistical quantities, compute layers—such as dense and convolutional layers—by default expect second raw moments as inputs and produce variances as outputs.
Conversely, activation functions expect variances as inputs and produce second raw moments.  
For architectures composed solely of compute layers and activation functions, this default behavior remains consistent.  
However, when additional layers are introduced, such as Max Pooling layers that both accept and produce variances (e.g., in LeNet-5), the activation function’s output and the compute layer’s input must be converted to variances.  

A special case arises in the \emph{first layer of the network}, where activation uncertainty—and thus variance—is unavailable.  
In this scenario, Equations~\ref{eq:pfp_dense_scalar:mean} and~\ref{eq:pfp_dense_scalar:E:variance} simplify to:  
\begin{equation}  
	\mu_{a_i^l} = \sum_{j=1}^{d_{l-1}} \mu_{w^l_{ij}} \cdot \mu_{x_j^{l-1}} \text{     and     } \sigma_{a_i^l}^2 = \sum_{j=1}^{d_{l-1}} \sigma_{w_{ij}^l}^2 \cdot \mu_{x_j^{l-1}}^2. \label{eq:pfp_dense_scalar:det_input}
\end{equation}  
These equations depend on weight variances, requiring the first layer's weight variances to be stored in variance format.  
For all subsequent compute layers, weight variance information can be stored directly as second raw moments to reduce computational overhead.  
Additionally, compute layers support three bias configurations: no bias, deterministic bias, and probabilistic bias with variances.  

This section analyzed the integration of custom operators into TVM, emphasizing design considerations specific to PFP operators.  
The evaluation of different implementation strategies demonstrated that the joint operator approach, combined with the second raw moment formulation, provides the highest efficiency in practice.

%% file: 06-tuning.tex
\section{Optimizing For Performance}
\label{sec:tuning}

\NewEnviron{scaletikzpicturetowidth}[1]{%
  \def\tikz@width{#1}%
  \def\tikzscale{1}\begin{lrbox}{\measure@tikzpicture}%
  \BODY
  \end{lrbox}%
  \pgfmathparse{#1/\wd\measure@tikzpicture}%
  \edef\tikzscale{\pgfmathresult}%
  \BODY
}

\newif\ifexternalfigures
\externalfiguresfalse

\begin{figure}
    \centering

	\begin{diffbox}

	\ifexternalfigures
		\includegraphics[width=\textwidth]{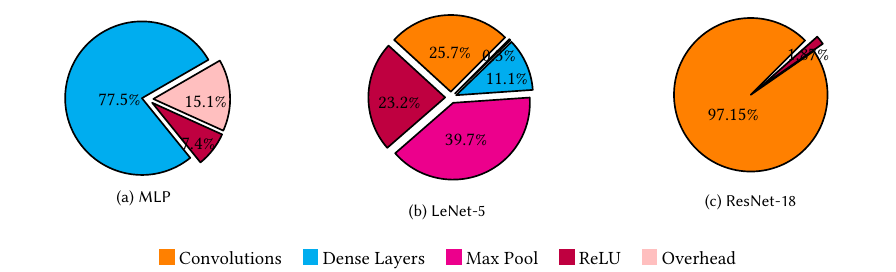}
    \else

		\begin{subfigure}[t]{0.32\textwidth}
			\centering
			\adjustbox{valign=t}{
				\begin{tikzpicture}[scale=0.65]
					\pie[sum=100, text=inside, radius=2, explode=0.15, rotate=30, color={cyan, purple, pink}]
					{77.5/\ , 7.4/\ , 15.1/\ }
				\end{tikzpicture}
			}
			\caption{MLP}
			\label{fig:tune:profiling_pie:mlp}
		\end{subfigure}
		\hfill
		\begin{subfigure}[t]{0.32\textwidth}
			\centering
			\adjustbox{valign=t}{
				\begin{tikzpicture}[scale=0.65]
					\pie[sum=100, text=inside, radius=2, explode=0.15, rotate=45, color={orange, purple, magenta, cyan, pink}]
					{25.7/\ , 23.2/\ , 39.7/\ , 11.1/\ } 
				\end{tikzpicture}
			}
			\caption{LeNet-5}
			\label{fig:tune:profiling_pie:lenet}
		\end{subfigure}
		\hfill
		\begin{subfigure}[t]{0.32\textwidth}
			\centering
			\adjustbox{valign=t}{
				\begin{tikzpicture}[scale=0.65]
					\pie[sum=100, text=inside, radius=2, explode=0.15, rotate=45, color={orange, purple}]
					{97.15/\ , 1.87/\ }
				\end{tikzpicture}
			}
			\caption{ResNet-18}
			\label{fig:tune:profiling_pie:resnet}
		\end{subfigure}

		\vspace{1em}
		\begin{center}
			\begin{tabular}{ccccc}
				\tikz\draw[orange,fill=orange] (0,0) rectangle (0.25,0.25); Convolutions &
				\tikz\draw[cyan,fill=cyan] (0,0) rectangle (0.25,0.25); Dense Layers &
				\tikz\draw[magenta,fill=magenta] (0,0) rectangle (0.25,0.25); Max Pool &
				\tikz\draw[purple,fill=purple] (0,0) rectangle (0.25,0.25); ReLU &
				\tikz\draw[pink,fill=pink] (0,0) rectangle (0.25,0.25); Overhead \\
			\end{tabular}
		\end{center}

	\fi

	\end{diffbox}
	
	\caption{Execution time distribution across operator types for three PFP-BNN architectures, measured on a Cortex-A72 with a mini-batch size of 10. While dense layers dominate latency in the MLP and convolutions in ResNet-18, LeNet-5 shows a more balanced distribution. Notably, otherwise trivial operators such as ReLU and Max Pool exhibit increased computational cost. The tooling and profiling overhead is typically negligible but becomes noticeable in the low-latency MLP.}
    \label{fig:tune:profiling_pie:bs10}
\end{figure}
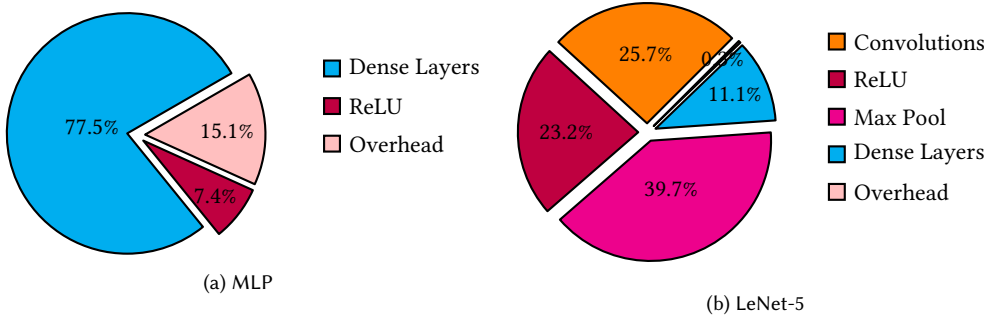

The PFP implementation presented in Section~\ref{sec:operator_library} provides a functional BNN framework capable of uncertainty prediction, as described in Section~\ref{sec:pfp:training}, and is already significantly faster than sampling-based BNNs.  
However, implementation- and hardware-specific safe optimizations can further enhance performance.  
First, profiling techniques identify the most computationally expensive operators, followed by optimization strategies and hand-tuned scheduling for the MLP, while additionally leveraging TVM-based automatic tuning frameworks~\cite{Chen2018tvm,Zheng2020ansor,Wu2023autotuning,Shao2022tvmMetaScheduler}.
Finally, we compare PFP implementations across various ARM processor architectures and against SVI-based BNNs.  

\subsection{Profiling Operators}
Efficient optimization requires focusing on the most computationally expensive operators, necessitating a detailed analysis of execution costs per operator.  
TVM provides three execution modes for compiled binaries: normal execution, benchmarking with advanced averaging for precise measurements, and profiling, which reports per-operator latencies.  
These detailed latency reports enable the evaluation of optimization effects on specific operators, as shown in Table~\ref{tab:tune:profiling:bs10}, and help visualize time distribution per operator, as illustrated in Figure~\ref{fig:tune:profiling_pie:bs10}.  

Figure~\ref{fig:tune:profiling_pie:bs10} illustrates the proportion of execution time spent per operator type.  
For the MLP, dense layers dominate computational costs, whereas in LeNet-5, latency is more evenly distributed across different layer types.  
This indicates that operators considered simple in a deterministic setting, such as activation functions or pooling layers, can become computationally complex when operating on distributions.  
This observation is further supported by detailed per-layer latency benchmarks and profiling results obtained with TVM on an ARM Cortex-A72, as shown in Table~\ref{tab:tune:profiling:bs10}.  

\subsection{Manual Optimizations}

\begin{DIFnomarkup}
\begin{table}
	\centering
	\small
	\caption{Comparison of manual optimization techniques for the PFP dense operator. Measurements on a Cortex-A72, with a 3-layer MLP and a mini-batch size of 10.}
	\begin{threeparttable}
	\begin{tabular}{c|c|r|r|r}
		\multicolumn{2}{c|}{\textbf{Optimizations}}														& \multicolumn{2}{c|}{\textbf{Latency}}						& \textbf{Speedup}					\\ \hline
		\textbf{Name}													& \textbf{Other Opt.}			& \textbf{without Opt.}	& \textbf{with Opt.}				& 									\\ \hline 
		Baseline (no tuning)											& OFF							& 3.760\,ms 			& - 								& - 								\\ 
		Baseline (min. tuning) 											& OFF							& 3.681\,ms				& -									& - 								\\ 
		\hline
		Tiling\tnote{1}													& OFF 							& 3.672\,ms				& 0.747\,ms							& 4.91$\times$							\\ 
		Loop Reordering													& OFF 							& 3.681\,ms				& 1.940\,ms							& 1.90$\times$							\\ 
		Vectorization													& OFF 							& 3.681\,ms				& 8.837\,ms							& 0.42$\times$						\\ 
		Parallelization													& OFF 							& 3.681\,ms				& 0.729\,ms							& 5.05$\times$							\\ 
		Loop Unrolling													& OFF 							& 3.681\,ms				& 1.967\,ms							& 1.87$\times$							\\
		\hline
		Tiling\tnote{1}													& ON 							& 0.754\,ms				& 4.237\,ms							& 0.18$\times$							\\ 
		Loop Reordering													& ON\tnote{2}  					& 0.750\,ms				& 0.743\,ms							& 1.01$\times$							\\ 
		Vectorization													& ON\tnote{2}  					& 0.759\,ms				& 0.743\,ms							& 1.02$\times$							\\ 
		Parallelization													& ON\tnote{2}  					& 1.953\,ms				& 0.743\,ms							& 2.63$\times$							\\ 
		Loop Unrolling													& ON\tnote{2}  					& 3.042\,ms				& 0.743\,ms							& 4.09$\times$							\\ 
		\hline
		All Optimizations 												& ON\tnote{2}  					& 3.760\,ms				& 0.743\,ms 						& 5.06$\times$						\\ 
	\end{tabular}
	\begin{tablenotes}
		\item[1] Without stochastic tuning.
		\item[2] All optimizations in use except tiling.
	\end{tablenotes}
	\end{threeparttable}
	\label{tab:dense_optimizations:bs10}
\end{table}
\end{DIFnomarkup}

The MLP is primarily constrained by the latency of dense layers, making the PFP dense operator the most promising target for optimization.  
We apply the following techniques, commonly used to accelerate matrix-matrix multiplication, to enhance the efficiency of probabilistic dense operators:  
\begin{itemize}  
	\item \emph{Tiling}: Partitioning matrices into smaller tiles to optimize memory access patterns. Tile sizes must be tuned.  
	\item \emph{Loop Reordering}: Adjusting loop order to improve vectorization, parallelization, and memory access efficiency.  
	\item \emph{Loop Unrolling}: Unrolling smaller loops to enhance vectorization and reduce loop overhead.  
	\item \emph{Vectorization}: Utilizing SIMD instructions efficiently, requiring careful selection of the appropriate dimension.  
	\item \emph{Parallelization}: Distributing computations across multiple cores while minimizing synchronization and communication overhead.  
\end{itemize}  

\begin{DIFnomarkup}
\begin{table} 
	\centering
	\small
	\caption{Comparison of Max Pool implementations. Measurements on a Cortex-A72 with a mini-batch size of 10.}
	\begin{tabular}{c|c|c|r|r}
		\textbf{Architecture}  					& \textbf{Implementation}						& \textbf{Auto-tuning}					& \multicolumn{2}{c}{\textbf{Latency}}								 	\\ \hline 
												&												&										& \textbf{Max Pools}				& \textbf{Entire Network}		 	\\ \hline
		LeNet-5 								& Generic Max Pool								& No									& 12.09\,ms							&	29.13\,ms					 	\\ 
		LeNet-5									& Generic Max Pool								& All operators							&  5.04\,ms							&	10.74\,ms					 	\\ 
		LeNet-5									& Generic Max Pool								& All except Max Pool					& 11.92\,ms							&	17.82\,ms					 	\\ 
		\hline
		LeNet-5 								& Vect. Max Pool $k=2$							& No									&  3.54\,ms							&	21.10\,ms					 	\\ 
		LeNet-5 								& Vect. Max Pool $k=2$							& All operators							& 27.28\,ms							&	33.42\,ms					 	\\ 
		LeNet-5 								& Vect. Max Pool $k=2$							& All except Max Pool					&  3.69\,ms							&	 9.79\,ms					 	
	\end{tabular}
	\label{tab:tune:maxpool:bs10}
\end{table}
\end{DIFnomarkup}

Table~\ref{tab:dense_optimizations:bs10} compares the impact of different optimization techniques, evaluating cases where only one optimization is active and where all optimizations except tiling are applied.  
This allows for an isolated analysis of each technique’s effectiveness.  
For instance, while vectorization is generally a promising optimization, it significantly degrades performance when applied alone, as it relies on a vectorizable inner loop, which must first be established through loop reordering.  

Tiling requires a separate evaluation.  
When applied independently with hand-tuned tile sizes, it proves highly effective.  
However, it is the only optimization that does not support stochastic tuning.  
Since the other optimizations benefit from stochastic tuning, enabling tiling disables this option.  
As a result, applying all optimizations, including tiling but without stochastic tuning, performs worse than basic tiling alone.  
The best performance is achieved by combining all other optimizations with stochastic tuning while excluding tiling.  
Thus, the effectiveness of optimizations is assessed without explicit tiling.  

\begin{DIFnomarkup}
\begin{table} 
	\centering
	\small
	\caption{Profiling of PFP neural architectures to determine most costly layers on a Cortex-A72 with a mini-batch size of 10.}
	\begin{threeparttable}
	\begin{tabular}{c|c|r|r|r|r|r}
		\textbf{Architecture}		& \textbf{Layer}		   	& \multicolumn{2}{c|}{\textbf{Baseline}} 	& \multicolumn{2}{c|}{\textbf{Tuned Implementation}} 						& \textbf{Speedup}			\\ \hline 
									&						   	& \textbf{Latency}					& \textbf{Fraction} 			& \textbf{Latency}			& \textbf{Fraction} 	&							\\ \hline 
		\multirow{5}*{MLP}			& Dense 1				   	&  2.931\,ms						&  62.8\,\%						&  0.642\,ms				&  33.7\,\%				&  4.6$\times$		\\ 
									& Dense 2				   	&  0.570\,ms						&  12.2\,\%						&  0.125\,ms				&   6.6\,\%				&  4.6$\times$		\\ 
									& ReLU\tnote{1}			   	&  0.195\,ms						&   4.2\,\%						&  0.185\,ms				&   9.7\,\%				&  1.1$\times$		\\ 
									& Dense 3				   	&  0.063\,ms						&   1.3\,\%						&  0.036\,ms				&   1.9\,\%				&  1.8$\times$		\\ 
									& Sum 						&  3.846\,ms						&  82.4\,\%						&  1.078\,ms				&  56.5\,\%				&  3.6$\times$		\\ 
									& Entire Network		  	&  4.668\,ms						& 100.0\,\%						&  1.908\,ms				& 100.0\,\%				&  2.4$\times$		\\ 
		\hline                                                                                                                                                                                
		\multirow{13}*{LeNet-5}	 	& Conv2d 2				   	&  7.207\,ms					 	&  30.4\,\%						&  1.509\,ms				&  12.4\,\% 			&  4.8$\times$		\\ 
								 	& ReLU 1				   	&  4.133\,ms						&  17.4\,\%						&  2.153\,ms				&  17.7\,\% 			&  1.9$\times$		\\ 
								 	& Dense 1				   	&  2.791\,ms						&  11.8\,\%						&  0.516\,ms				&   4.2\,\% 			&  5.4$\times$		\\ 
									& Max Pool 1\tnote{1,2}	   	&  2.780\,ms						&  11.7\,\%						&  2.807\,ms				&  23.1\,\% 			&  1.0$\times$		\\ 
								 	& ReLU 2				   	&  1.541\,ms						&   6.5\,\%						&  0.980\,ms				&   8.1\,\% 			&  1.6$\times$		\\ 
								 	& Conv2d 1				   	&  0.949\,ms						&   4.0\,\%						&  0.552\,ms				&   4.5\,\% 			&  1.7$\times$		\\ 
									& Max Pool 2\tnote{2}	   	&  0.854\,ms						&   3.6\,\%						&  0.902\,ms				&   7.4\,\% 			&  0.9$\times$		\\ 
								 	& Dense 2				   	&  0.589\,ms						&   2.5\,\%						&  0.111\,ms				&   0.9\,\% 			&  5.3$\times$		\\ 
								 	& ReLU 3				   	&  0.168\,ms						&   0.7\,\%						&  0.071\,ms				&   0.6\,\% 			&  2.4$\times$		\\ 
								 	& ReLU 4				   	&  0.107\,ms						&   0.5\,\%						&  0.051\,ms				&   0.4\,\% 			&  2.1$\times$		\\ 
								 	& Dense 3				   	&  0.057\,ms						&   0.2\,\%						&  0.027\,ms				&   0.2\,\% 			&  2.1$\times$		\\ 
									& Sum 		 				& 21.751\,ms						&  91.8\,\%						& 10.226\,ms				&  84.1\,\% 			&  2.1$\times$		\\ 
									& Entire Network		   	& 23.698\,ms						& 100.0\,\%						& 12.166\,ms				& 100.0\,\% 			&  1.9$\times$		\\ 

	\end{tabular}
	\begin{tablenotes}
		\item[1] Layers present multiple times in network
		\item[2] Layers excluded from tuning
	\end{tablenotes}
	\end{threeparttable}
	\label{tab:tune:profiling:bs10}
\end{table}
\end{DIFnomarkup}

Loop unrolling and parallelization emerge as the most critical optimizations for the PFP dense operator.  
Combining these techniques results in a substantial speedup of $5\times$ compared to the untuned network.  

\paragraph{Max Pool Operator}
For LeNet-5, the Max Pooling operator by \citeauthor{roth2021phd}~\cite{roth2021phd} is formulated as a generic reduction, which is suboptimal in performance.  
We implement a specialized, vectorized Max Pool operator with fixed kernel size for improved efficiency.
As evidenced in Table~\ref{tab:tune:maxpool:bs10}, the application of automatically generated schedules fails to enhance performance and, in fact, results in a substantial deterioration in runtime efficiency.
Consequently, this hand-optimized operator is excluded from further automatic tuning.

\subsection{Automatic Optimizations}
While implementing custom schedules for operators remains common, significant advancements in automatic tuning frameworks have been made by the TVM community~\cite{Chen2018tvm,Zheng2020ansor,Wu2023autotuning,Shao2022tvmMetaScheduler}.  
The \emph{Meta Scheduler}~\cite{Shao2022tvmMetaScheduler} automates schedule generation, eliminating the need for manual implementations.  
Conceptually, the Meta Scheduler is a domain-specific probabilistic programming abstraction that explores a large optimization search space.  
Auto-tuning then navigates this space, benchmarking different implementation variants on the target hardware.  
While this process is slower than expert-crafted schedules, it typically achieves comparable performance and does not require manual effort from domain experts.
Applying all optimizations from Table~\ref{tab:dense_optimizations:bs10}, we achieve nearly identical latencies: $0.743\,ms$ with handwritten schedules and $0.742\,ms$ with the Meta Scheduler.  
Thus, the Meta Scheduler proves highly effective, even for specialized PFP operators, and is used for further experiments.  

Table~\ref{tab:tune:profiling:bs10} presents profiling results for the PFP MLP and LeNet-5 before and after tuning, highlighting operator latencies.  
All operators, except Max Pool layers, benefit significantly from tuning, with dense and convolution layers showing the most substantial acceleration.  

\subsection{Evaluation and Performance Gain}

\begin{figure}[t]
    \centering
	\begin{subfigure}{0.40\textwidth} 
    	\centering
    	\includegraphics[width=\textwidth]{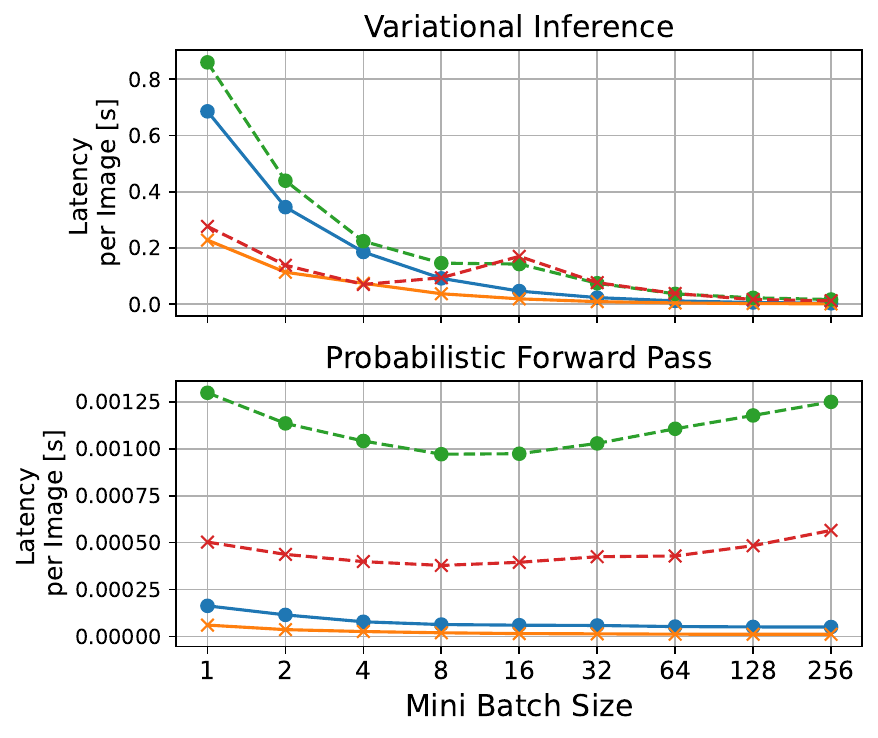}    
		\caption{Latency per Image}
	\end{subfigure}
	\begin{subfigure}{0.58\textwidth} 
    	\centering
    	\includegraphics[width=\textwidth]{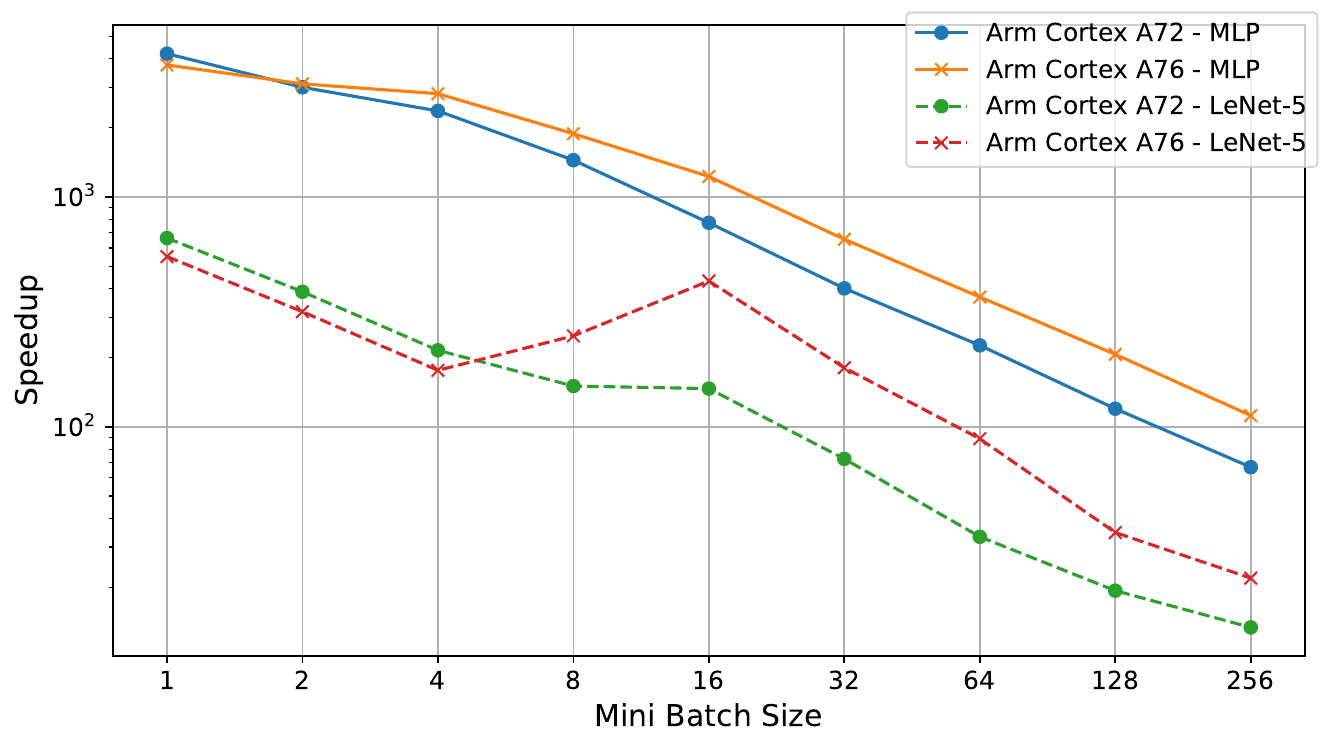}    
		\caption{Relative Speedup of PFP Over SVI}
    	\label{fig:tune:speedup_and_latency_batchsize:speedup}
	\end{subfigure}
\caption{Speedup and latency in relation to mini-batch size. The SVI-BNN, evaluated using 30 samples, exhibits high latency and poor scalability for small mini-batch sizes, resulting in significantly higher per-image latency. In contrast, the PFP implementation maintains consistent performance across all mini-batch sizes due to targeted tuning. The observed speedup highlights the advantages of PFP, particularly for small mini-batch sizes, which are critical for low-latency processing in embedded systems.}
    \label{fig:tune:speedup_and_latency_batchsize}
\end{figure}

In resource-constrained embedded systems, real-time and low-latency requirements often necessitate the use of small mini-batch sizes.
Figure~\ref{fig:tune:speedup_and_latency_batchsize} shows that SVI-based BNNs scale poorly with decreasing mini-batch sizes, incurring substantial computational overhead and high latency per image.  
To reflect realistic constraints where minimal sample counts are preferred, the SVI-BNN---implemented using Pyro---is evaluated with only 30 samples, which is at the lower bound of the range (see Figure~\ref{fig:number_of_samples_vi}).
In contrast, the PFP implementation is optimized per mini-batch size, maintaining relatively stable latency.  
Only minor latency increases occur when mini-batch sizes misalign with cache or SIMD instruction sizes.  

Overall, the performance gap is substantial, reaching multiple orders of magnitude, as illustrated by the speedups in Figure~\ref{fig:tune:speedup_and_latency_batchsize:speedup}.
For a mini-batch size of 256, speedups range from $13\times$ to $112\times$, while for mini-batch size 1—common in embedded systems—they increase dramatically, reaching $38\times$ to $3936\times$.  

\begin{DIFnomarkup}
\begin{table}
	\centering
	\small
    \begin{diffbox}
	\begin{threeparttable}
	\caption{Algorithm Performance Comparison on Embedded ARM Processors, using vectorized Max Pool for Dirty-MNIST.}
	\begin{tabular}{c|c|c|r|r|r|r|r|r}
		\textbf{Architecture}   	& \textbf{batch-size} 	& \textbf{Processor} 				& \multicolumn{2}{c|}{\textbf{Deterministic NN}}			& \multicolumn{1}{c|}{\textbf{SVI}}	& \multicolumn{2}{c|}{\textbf{PFP}}				& \textbf{Speedup} 	\\ \hline
		  						&							& 									& \textbf{not tuned} 	& \textbf{tuned} 					&  						& \textbf{not tuned}	& \textbf{tuned} 							\\ \hline							
		\multirow{3}*{MLP}		&	\multirow{3}*{1}		&	Cortex-A53			 			&   1.49\,ms			&  0.396\,ms						& 		-				&  1.53\,ms				&   0.548\,ms			&	 				\\ 
								&							&	Cortex-A72			 			&   0.49\,ms			&  0.094\,ms						&  	598.25\,ms			&  0.42\,ms				&   0.152\,ms			&	 3935.9$\times$	\\ 
								&							&	Cortex-A76			 			&   0.18\,ms			&  0.044\,ms						&  	226.76\,ms			&  0.20\,ms				&   0.062\,ms			&	 3657.4$\times$	\\ 
		\hline                                                                                                                                                                                                              		            
	
		\multirow{3}*{MLP}		&	\multirow{3}*{10}		&	Cortex-A53			 			&  14.02\,ms			&  0.933\,ms						& 		-				&  15.26\,ms			&   4.989\,ms			&	  				\\ 
								&							&	Cortex-A72			 			&   4.59\,ms			&  0.186\,ms						&   734.74\,ms			&   3.75\,ms			&   0.742\,ms			&	 990.2$\times$	\\ 
								&							&	Cortex-A76			 			&   1.64\,ms			&  0.071\,ms						&   307.52\,ms			&   1.89\,ms			&   0.341\,ms			&	 901.8$\times$	\\ 
		\hline                                                                                                                                                                                                              		            
		\multirow{3}*{MLP}		&	\multirow{3}*{100}	   	&	Cortex-A53			 			& 137.80\,ms			&  6.565\,ms						& 		-				& 147.61\,ms			&  15.358\,ms			&	      			\\ 
								&						   	&	Cortex-A72			 			&  45.81\,ms			&  1.134\,ms						&   775.32\,ms			&  36.33\,ms			&   5.182\,ms			&	 149.6$\times$	\\ 
								&						   	&	Cortex-A76			 			&  16.30\,ms			&  0.230\,ms						&   306.89\,ms			&  18.60\,ms			&   1.200\,ms			&	 255.7$\times$	\\ 
		\hline                   				                                                                                                                                                                            		            
		\hline                   				                                                                                                                                                                            		            
		\multirow{3}*{LeNet-5}	&	\multirow{3}*{1}		&	Cortex-A53			 			&   2.61\,ms			&  0.875\,ms						& 		-				&  7.98\,ms				&   4.396\,ms			&	      		\\ 
								&							&	Cortex-A72			 			&   0.84\,ms			&  0.242\,ms						& 	743.52\,ms			&  2.23\,ms				&   1.377\,ms			&	 540.0$\times$	\\ 
								&							&	Cortex-A76			 			&   0.36\,ms			&  0.128\,ms						& 	272.60\,ms			&  0.92\,ms				&   0.502\,ms			&	 543.0$\times$	\\ 
		
		\hline                   				                                                                                                                                                                            		            
		\multirow{3}*{LeNet-5}	&	\multirow{3}*{10}		&	Cortex-A53			 			&  21.14\,ms			&  4.726\,ms						& 		-				&  76.09\,ms			&  35.159\,ms			&	      			\\ 
								&							&	Cortex-A72			 			&   6.89\,ms			&  0.754\,ms						&  1196.42\,ms			&  21.23\,ms			&  10.022\,ms			&	 119.4$\times$	\\ 
								&							&	Cortex-A76			 			&   3.16\,ms			&  0.347\,ms						&   801.40\,ms			&   9.63\,ms			&   3.897\,ms			&	 205.6$\times$	\\ 
		\hline                   				                                                                                                                                                                            		            
		\multirow{3}*{LeNet-5}	&	\multirow{3}*{100}		&	Cortex-A53			 			& 209.28\,ms			& 41.697\,ms						& 		-				& 801.33\,ms			& 383.680\,ms			&	      			\\ 
								&							&	Cortex-A72			 			&  70.08\,ms			&  9.524\,ms						&  2708.16\,ms			& 240.94\,ms			& 116.330\,ms			&	  23.3$\times$	\\ 
								&							&	Cortex-A76			 			&  31.51\,ms			&  3.131\,ms						&  2488.73\,ms			& 119.76\,ms			&  45.039\,ms			&	  55.3$\times$	\\ 
		\hline                   				                                                                                                                                                                            		            
		\hline                   				                                                                                                                                                                            		            
		\multirow{3}*{ResNet-18} & 	\multirow{3}*{1}		&	Cortex-A53			 			&   2.70\,s				&  0.134\,s							& 		-				&   6.22\,s				&   1.981\,s			&	  				\\ 
								&							&	Cortex-A72			 			&   0.98\,s				&  0.035\,s							&   29.075\,s			&   2.04\,s				&   0.763\,s			&	 38.1$\times$	\\ 
								&							&	Cortex-A76			 			&   0.49\,s				&  0.013\,s							&   14.742\,s			&   0.75\,s				&   0.158\,s			&	 93.3$\times$	\\ 
		\hline                   				                                                                                                                                                                         		            
		\multirow{3}*{ResNet-18} & 	\multirow{3}*{10}		&	Cortex-A53			 			&  27.10\,s				&  1.354\,s	   						& 		-				&  66.39\,s				&  19.001\,s			&					\\ 
								&							&	Cortex-A72			 			&   9.92\,s				&  0.349\,s							&  	42.043\,s			&  18.70\,s				&   6.960\,s			&	  6.0$\times$ 	\\ 
								&							&	Cortex-A76			 			&   4.94\,s				&  0.126\,s							&  	21.385\,s			&   7.62\,s				&   1.784\,s			&	 12.0$\times$	\\ 
		\hline                   				                                                                                                                                                                          		            
		\multirow{3}*{ResNet-18} & 	\multirow{3}*{100}		&	Cortex-A53			 			& 270.79\,s				& 17.153\,s\tnote{$\dagger$}		& 		-				& 699.96\,s				& 203.450\,s\tnote{$\dagger$}&				\\ 
								&							&	Cortex-A72			 			&  99.33\,s				&  3.507\,s							&  155.095\,s			& 207.39\,s				&  69.068\,s			&	 2.2$\times$	\\ 
								&							&	Cortex-A76			 			&  49.79\,s				&  1.365\,s							&   77.221\,s			&  76.09\,s				&  17.425\,s			&	 4.4$\times$
	\label{tab:performance}
	\end{tabular}

	\begin{tablenotes}
	\footnotesize
	\item[$\dagger$] Meta Scheduler optimization was limited to 200 iterations due to the substantial runtime required for the full 2000-iteration schedule.
	\end{tablenotes}

	\end{threeparttable}
	\end{diffbox}
\end{table}
\end{DIFnomarkup}

Table~\ref{tab:performance} compares probabilistic SVI-based BNNs and PFP-based architectures across multiple embedded ARM processors.  
Auto-tuning yields clear benefits for both deterministic and PFP implementations.  

On average, PFP is $3.4\times$ slower for the MLP, $9.2\times$ slower for LeNet-5 and $15.7\times$ slower for ResNet-18 compared to their deterministic counterparts.  
This slowdown is expected due to the increased computational complexity and the doubling of both parameters and activations.  
However, compared to a state-of-the-art SVI-based BNN with only 30 samples, PFP achieves substantial average speedups of $1648\times$ for the MLP, $248\times$ for LeNet-5 and $26\times$ for the ResNet-18.  

These results highlight the efficiency gains of an approximate SVI approach combined with code generation, demonstrating the feasibility of deploying such models on resource-constrained embedded systems.

%% file: 06a-pBNNs.tex
\section{Accelerating PFP Networks through Partiality}
\label{sec:partial}

As shown in prior work, partial BNNs can retain strong predictive quality while reducing computational complexity~\cite{sharma2023partialBnns,steger2024rlleWorkshop,Andrade2025}.
However, the combination of partial BNNs and PFP remains unexplored.
We therefore evaluate full and last-layer probabilistic ResNet-18 variants on CIFAR-10 and CINIC-10, chosen due to their comparatively high task complexity.
Last-layer BNNs are particularly attractive from a performance perspective, as they promise substantially lower inference cost.
For the SVI baselines, all models were trained using the same hyperparameters as in Section~\ref{sec:pfp:training}.

Table~\ref{tab:resnet18_probabilistic_performance_a72} shows that SVI accuracy is largely unaffected by partiality, whereas OOD detection performance depends more strongly on the amount of stochasticity.
For CIFAR-10, both full and last-layer SVI-BNNs achieve comparable AUROC, while on the more challenging CINIC-10 benchmark the fully probabilistic model performs better.
Transferred and calibrated PFP networks preserve accuracy reasonably well on CIFAR-10, but exhibit reduced AUROC for last-layer configurations.
On CINIC-10, both accuracy and AUROC degrade substantially compared to SVI, particularly for partial BNNs.
Fine-tuning consistently improves PFP OOD-detection performance and, on CIFAR-10, partially probabilistic PFP even surpasses the SVI baseline in AUROC.
However, the gap to SVI remains substantial on CINIC-10, suggesting that reduced stochastic capacity and the Gaussian activation approximation become increasingly limiting for more complex tasks.

From a computational perspective, last-layer PFP-BNNs substantially outperform fully probabilistic variants.
While untuned inference achieves roughly a 2$\times$ speedup, TVM tuning increases this advantage to up to 28$\times$.

Overall, combining partial BNNs with PFP offers significant computational benefits, but the resulting uncertainty quality remains strongly task-dependent.

\begin{DIFnomarkup}
\begin{table}[tb]
    \centering
    \small
    \caption{Quality and PFP inference latency of ResNet-18 variants with probabilistic partiality (Cortex-A72, mini-batch size of 1)}
    \label{tab:resnet18_probabilistic_performance_a72}

	\begin{diffbox}
    \begin{tabular}{l l | cc | ccc | cc | rr}
        \hline
        \textbf{Dataset}
        & \textbf{Prob.}
        & \multicolumn{2}{c|}{\textbf{SVI}}
        & \multicolumn{3}{c|}{\textbf{PFP Calibrated}}
        & \multicolumn{2}{c|}{\textbf{PFP Fine-tuned}}
        & \multicolumn{2}{c}{\textbf{PFP Latency}} \\
        \cline{3-4}\cline{5-7}\cline{8-9}\cline{10-11}

        & \textbf{Layers}
        & \textbf{Acc. }
        & \textbf{AUROC}
        & \textbf{Calib.}
        & \textbf{Acc. }
        & \textbf{AUROC}
        & \textbf{Acc. }
        & \textbf{AUROC}
        & \textbf{Not tuned}
        & \textbf{Tuned} \\
        \hline

        \multirow{2}{*}{CIFAR-10}
            & All       & 91.5\% & 89.2\% & 0.40 & 91.5\% & 94.1\% & 91.2\% & 95.9\% & 2366.3\,ms & 1024.7\,ms \\ 
            & Last-only & 92.5\% & 89.0\% & 0.05 & 89.3\% & 86.9\% & 92.0\% & 95.6\% & 1207.1\,ms &   36.1\,ms \\ 
        \hline

        \multirow{2}{*}{CINIC-10}
            & All       & 81.5\% & 86.2\% & 0.45 & 62.7\% & 75.7\% & 63.9\% & 78.3\% & 2366.3\,ms & 1024.7\,ms \\ 
            & Last-only & 82.4\% & 81.0\% & 0.05 & 56.0\% & 50.3\% & 37.7\% & 60.2\% & 1207.1\,ms &   36.1\,ms 

    \end{tabular}
	\end{diffbox}
\end{table}
\end{DIFnomarkup}

%% file: 07-conclusion.tex
\section{Conclusion}
\label{sec:conclusion}

Bayesian neural networks empower neural networks with probabilistic reasoning, enabling them to issue warnings when confidence in a prediction is low.
Distinguishing between aleatoric and epistemic uncertainty allows identification of inherent stochasticity versus uncertainty due to insufficient training data.  
In safety-critical applications, such as transportation systems, embedding BNNs in devices enables them to trigger warnings under high uncertainty, allowing human operators to intervene when needed.  
However, despite these advantages, BNNs are rarely deployed in resource-constrained embedded systems due to their substantial computational and memory overhead compared to non-probabilistic models.  

This work advances the deployment of probabilistic machine learning models on embedded systems.  
We demonstrate how Stochastic Variational Inference-based BNNs can be efficiently executed on embedded ARM CPUs by leveraging the Probabilistic Forward Pass approximation.  
To achieve this, we extend TVM, a deep learning compiler, to support essential probabilistic operators and optimize them for target hardware architectures.

The Probabilistic Forward Pass mitigates the primary computational bottleneck of BNNs---the need for multiple forward passes per prediction---by assuming Gaussian-distributed weights and activations, enabling a single analytical forward pass.  
This approach extends the core assumption of SVI-based BNNs, which model weights as parametric distributions, to the activation domain.  
While this limits the network’s ability to capture complex, non-Gaussian activation distributions, it significantly reduces computational costs when Gaussian approximations are sufficient.  

A key advantage of PFP is its compatibility with SVI-trained BNNs.
We show that direct conversion already achieves higher performance than standard PFP training.
However, our newly introduced staged SVI-to-PFP training methodology enhances PFP performance even further.
We demonstrate that this methodology successfully achieves uncertainty disentanglement quality on par with SVI, while simultaneously scaling to large networks and complex datasets.

Efficient hardware implementation is essential for deploying probabilistic models on embedded systems.  
Since they rely on specialized operators for Gaussian distributions—unsupported by standard vendor libraries for machine learning acceleration—we extend the deep learning compiler TVM.

During evaluation of multiple implementation variants we show that joint operators, which compute mean and variance in a single operation, yield greater efficiency.  
Further optimizations, such as eliminating unnecessary variance conversions and simplifying the first network operators for deterministic inputs, further reduce computational costs.  

Following manual optimizations, additional performance gains were achieved through profiling-based operator tuning.  
We applied targeted optimizations to probabilistic dense operators, developed a more efficient Max Pool operator, and leveraged TVM’s Meta Scheduler for auto-tuning.  
As a result, our optimized Probabilistic Forward Pass implementation achieved a two-order-of-magnitude speedup over the SVI-based BNN baseline on ARM processors.  
For small mini-batch sizes, which are critical for low-latency embedded applications, speedups of up to $3900\times$ were achieved.  
Finally, in a combined approximation study, we demonstrate that partial PFP networks can achieve comparable performance in some tasks, while achieving another $28\times$ speedup.
Of particular interest for future work is the evaluation of PFP-based models on larger and more diverse tasks, such as large-scale image classification on ImageNet or semantic segmentation.

To our knowledge, this work presents the first end-to-end demonstration of training, optimization, and deployment of Bayesian neural networks on resource-constrained embedded systems.
Our findings illustrate that integrating Bayesian approximations with deep learning compilers enables the deployment of otherwise computationally prohibitive probabilistic models.
We hope this work marks the beginning of further research bridging the gap between resource-intensive probabilistic models and constrained embedded hardware—bringing uncertainty-aware neural networks into everyday devices and enabling them to acknowledge uncertainty by saying, "I don't know."

%% file: acknowledgement.tex
\begin{DIFnomarkup}
\begin{acks}
The authors gratefully acknowledge the financial support under the scope of the COMET program within the K2 Center “Integrated Computational Material, Process and Product Engineering (IC-MPPE)” (Project No 886385). This program is supported by the Austrian Federal Ministries for Climate Action, Environment, Energy, Mobility, Innovation and Technology (BMK) and for Labour and Economy (BMAW), represented by the Austrian Research Promotion Agency (FFG), and the federal states of Styria, Upper Austria and Tyrol.

We further acknowledge the support by the state of Baden-Württemberg through bwHPC and the German Research Foundation
(DFG) through grant INST 35/1597-1 FUGG.
\end{acks}
\end{DIFnomarkup}